\documentclass{article} 
\usepackage{iclr2026_conference,times}
\usepackage{booktabs, multirow}
\usepackage{wrapfig}
\usepackage{subcaption}   
\usepackage{comment}
\usepackage{float} 
\usepackage[pagebackref=true,colorlinks,citecolor=brown]{hyperref} 

\usepackage{amsmath,amsfonts,bm}

\def\eqref#1{equation~\ref{#1}}

\def\1{\bm{1}}

\DeclareMathAlphabet{\mathsfit}{\encodingdefault}{\sfdefault}{m}{sl}
\SetMathAlphabet{\mathsfit}{bold}{\encodingdefault}{\sfdefault}{bx}{n}

\newtheorem{requirement}{Requirement}
\newcommand{\twnote}[1]%
    {\textcolor{cyan}{\textbf{TW: #1}}}

\usepackage{xcolor}
\newcommand{\blue}[1]{\textcolor{blue}{#1}}

\usepackage{hyperref}
\usepackage{url}

\usepackage{hang}
\usepackage{graphicx}
\title{Emergent Dexterity via Diverse Resets and Large-Scale Reinforcement Learning}
\newcommand{\Method}{\texttt{OmniReset}}
\iclrfinalcopy 

\author{
Patrick Yin$^{1,*}$ \quad
Tyler Westenbroek$^{1,*}$ \quad
Octi Zhang$^{2}$ \quad
Joshua Tran$^{1}$ \quad
Ignacio Dagnino$^{1}$
\And
Eeshani Shilamkar$^{1}$ \quad
Numfor Mbiziwo-Tiapo$^{1}$ \quad
Simran Bagaria$^{3}$ \quad
Xinlei Liu$^{1}$
\And
Galen Mullins$^{3}$ \quad
Andrey Kolobov$^{3}$ \quad
Abhishek Gupta$^{1}$ \\
$^{1}$University of Washington \quad
$^{2}$NVIDIA \quad
$^{3}$Microsoft Research \quad
$^{*}$Equal contribution
}

\begin{document}

\maketitle

\begin{figure}[h!]
    \centering

\includegraphics[width=\linewidth]{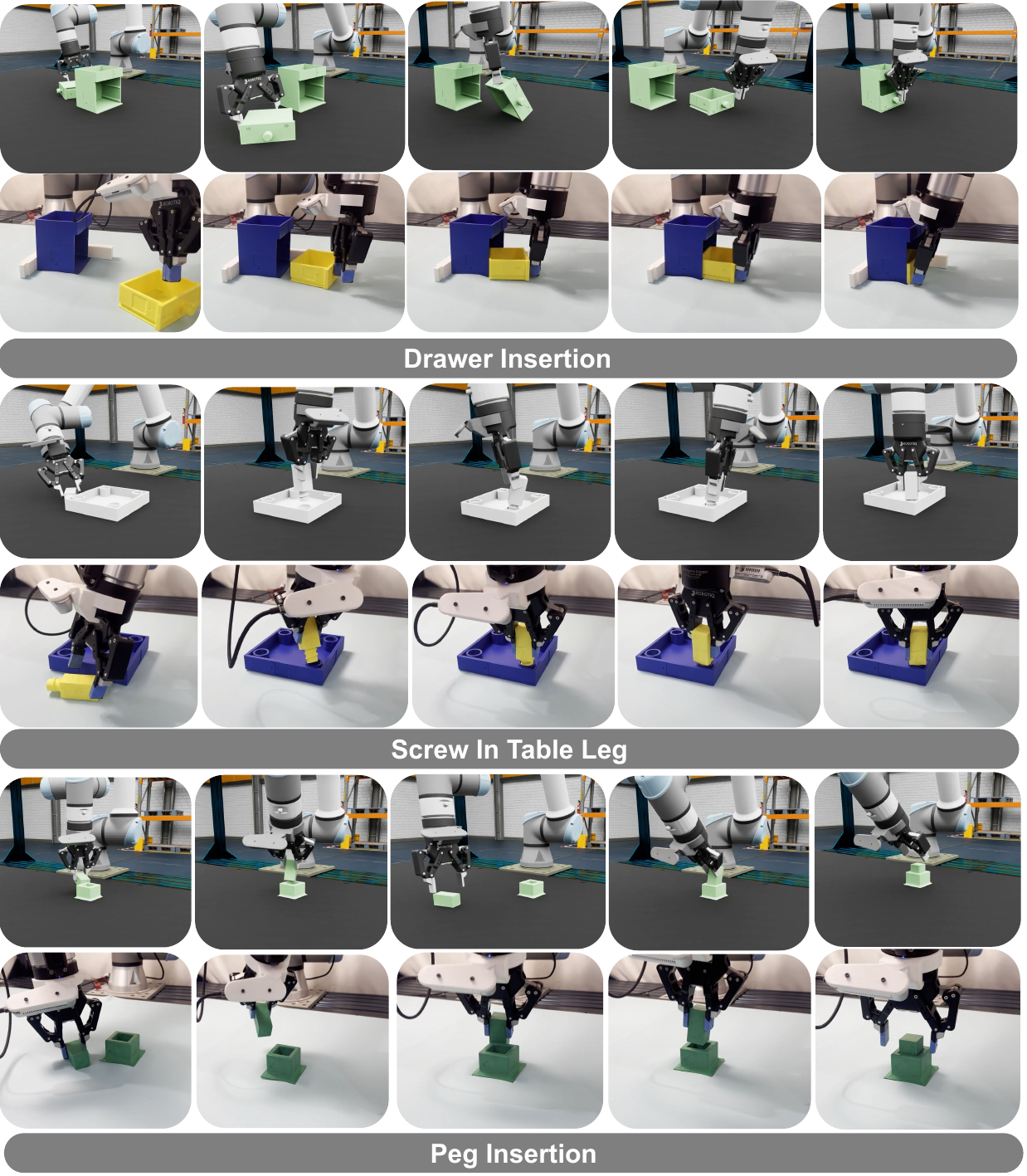}
    \caption{\footnotesize{\Method \ is a scalable framework for dexterous manipulation which uses diverse simulator resets and large scale reinforcement learning to solve contact-rich, long horizon tasks beyond the capabilities of existing approaches. These policies are then distilled to RGB camera observations and robustly transferred to the real world zero-shot, where they are able to consistently solve challenging tasks over much wider ranges of initial conditions than baselines.}}
    \label{fig:leg_sim_real}
\end{figure}

\begin{abstract}
Reinforcement learning in massively parallel physics simulations has driven major progress in sim-to-real robot learning. However, current approaches remain brittle and task-specific, relying on extensive per-task engineering to design rewards, curricula, and demonstrations. Even with this engineering, typical reinforcement learning methods can often fail on long-horizon, contact-rich manipulation tasks and do not meaningfully scale with compute, as performance quickly saturates when training revisits the same narrow regions of state space. We introduce \Method, a simple and scalable framework that enables on-policy reinforcement learning to robustly solve a broad class of dexterous manipulation tasks using fixed algorithm hyperparameters, no curricula, minimal reward engineering, and no human demonstrations. Our key insight is that long-horizon exploration can be dramatically simplified by using simulator resets to systematically expose the RL algorithm to the diverse set of robot-object interactions that underlie dexterous manipulation. \Method\ programmatically generates such resets with minimal human input, converting additional compute directly into broader behavioral coverage and continued performance gains for dynamic policies. We show that \Method\ gracefully scales to long-horizon dexterous manipulation tasks beyond the capabilities of existing approaches and is able to learn robust policies demonstrating a variety of dynamic, contact-rich recovery behavior. Finally, we distill \Method \ into visuomotor policies that can be transferred to the real world zero-shot, displaying robust retrying behavior to accomplish complex, contact-rich tasks with non-trivial success rates. Project webpage: https://omnireset.github.io
\end{abstract}

\section{Introduction}
Reinforcement learning (RL) in massively parallelized simulation environments \citep{mittal2023orbit, todorov2012mujoco} has driven recent successes in sim-to-real robotics \citep{akkaya2019solving, hwangbo2019learning}. While very successful for locomotion and navigation problems, these methods have seen relatively less success in robotic manipulation problems. In principle, these algorithms can also be applied to robotic manipulation, automatically acquiring complex, contact-rich behaviors through repeated interaction with the simulated environment. Yet in practice, obtaining robust and performant policies still requires extensive per-task environment and reward engineering, limiting the scalability of this paradigm—particularly for long-horizon manipulation tasks. This sharply contrasts with domains such as language modeling, where simply scaling data and compute has yielded dramatic gains with similarly simple RL algorithms \cite{guo2025deepseek}. How can we achieve comparable scalability for robotic manipulation? 

The central bottleneck in robotic RL is that standard exploration techniques \cite{schulman2017proximal, haarnoja2018soft} saturate as parallelism and compute are scaled, repeatedly sampling a narrow state–action distribution and becoming trapped in local minima \cite{singla2024sapg}. While numerous advanced exploration methods have been proposed \cite{pathak2017curiosity, burda2018exploration}, their added algorithmic complexity makes them difficult to scale in practice. As a result, practitioners often rely on human intuition to reduce the exploration burden, injecting structure through task-specific reward design \cite{westenbroek2022lyapunov, ng2000algorithms, handa2023dextreme}, hand-designed curricula \cite{lee2020learning}, or user-provided demonstrations \cite{bauza2025demostart, peng2018deepmimic, nair2018overcoming}. Although effective in many settings, these approaches are fundamentally limited by the amount of human effort they require. A complementary strategy is to simplify the learning problem through additional system scaffolding—using RL only for contact-rich phases while relying on motion planning or trajectory optimization for the remainder \cite{lee2020guided, tang2024automate, zhou24spire,lee20guapo}. While this reduces exploration demands, it comes at the cost of increased system complexity and a lack of smooth, adaptive behavior for complex tasks. These approaches reflect a prevailing assumption in robotics: that dexterous manipulation is too complex to emerge from large-scale RL alone and must instead be scaffolded with additional task-specific structure.

We argue that with the right system design, this scaffolding is unnecessary. In this work, we instead propose \emph{systematically exposing RL to a superset of the interactions it will encounter when manipulating the scene}, and then allowing dexterous task-specific behaviors to emerge from large-scale compute and optimization. Although the space of possible behaviors for robust task performance is vast—flipping, screwing, insertion, and other contact-rich motions—successful policies reuse a relatively small set of recurring interaction modes, such as approaching objects, initiating contact, and forming stable grasps. These modes can be densely covered through generic resets that do not encode task-specific solutions. By sufficiently randomizing object poses and sampling these interaction states, we substantially reduce the exploration burden and ensure the agent encounters meaningful success signals. This coverage allows sparse rewards to propagate smoothly through the state space, enabling the agent to identify high-value regions and learn how to stitch together multiple distinct behaviors to reach these goals, without task-specific reward shaping or guidance.

Concretely, we introduce \Method, a scalable framework for robotic manipulation that automatically generates diverse initial-state distributions to densely cover the contact-rich interactions the robot may encounter. This coverage allows PPO \cite{schulman2017proximal} to fully leverage large-scale compute, improving performance as the number of parallel environments increases. When scaled sufficiently, \Method\ learns the emergent behavior of combining multiple interaction modes—such as pushing, flipping, and insertion (Fig. \ref{fig:leg_sim_real})—into coherent, multi-stage strategies without task-specific reward shaping, curricula, or demonstrations. Across diverse contact-rich tasks, this approach solves long-horizon problems that are considerably out of reach for existing methods, producing robust policies that succeed from a huge range of initial states rather than narrow distributions typical of many prior approaches. Finally, we demonstrate that \Method \ can be used to train visuomotor policies via student-teacher distillation which can robustly transfer zero-shot to the real-world, substantially outperforming alternatives such as imitation learning.

\section{Related Work}
\textbf{Exploiting Resets in Reinforcement Learning}: Exploiting simulator resets for RL is a natural idea which has been explored in many contexts. Prior theoretical works \citep{kakade2002approximately} have suggested more uniform sampling over initial states, but do not provide practical algorithms. The primary focus of many works is to make learning more tractable by generating an explicit curriculum over resets \citep{tang2023industreal,dennis2020emergent, bauza2025demostart}, for instance through a ``reverse-curriculum" of states going backwards from the goal \citep{florensa2017reverse} or using a learned dynamics model to propose viable resets \citep{edwards2018forward, ivanovic2019barc}. In contrast, a second category of methods leverage \emph{demonstrations} (whether human or automatically generated) to generate feasible pathways to the goal \citep{tao2024reverse, resnick2018backplay, salimans2018learning, bauza2025demostart, tang2024automate}. In contrast to these prior works, we show that neither human demonstrations, nor a curriculum is needed, but rather that the simple approach for generating diverse resets in \Method \ naturally scales to various long-horizon manipulation problems without added algorithmic complexity. More recently, staggered environment resets \citep{bharthulwar2025staggeredenvironmentresetsimprove} have been shown to improve optimization by decorrelating rollouts, but leave the initial state distribution unchanged and thus do not alleviate exploration. In contrast, \Method \ directly addresses exploration by enumerating diverse, task-relevant initial states, enabling coverage of critical intermediate states that are otherwise rarely visited.

\textbf{Exploration Strategies for Reinforcement Learning}: RL practitioners have designed a variety of exploration strategies to effectively uncover goal-reaching paths for a \emph{fixed set of initial conditions}, with uninformative rewards. A major line of work is bonus-based exploration, where agents receive intrinsic rewards for visiting novel or unpredictable states. Count-based methods reward visits to rarely seen states \citep{ostrovski2017count,bellemare2016unifying,martin2017count}, while curiosity-based methods provide bonuses based on prediction errors~\citep{burda2018exploration,sancaktar2022curious,pathak2017curiosity}. Other approaches~\citep{osband2016deep, osband19rvf, russothompson} promote temporally correlated exploration by injecting stochasticity at the policy or value-function level. Finally, diversity-driven methods optimize for behavioral variety \citep{eysenbach2018diversity, rajeswarurl}. These approaches are complimentary to our work; our main contribution is demonstrating that large scale-scale parallelization and resetting schemes lead to the emergence of surprising levels of dexterity without the need for advanced exploration incentives. 

\textbf{Leveraging Demonstrations}: An alternative is to increasingly rely on human demonstrations and imitation learning to overcome difficult long-horizon exploration. Approaches include adding auxiliary BC loss terms to RL objectives \citep{nair2018overcoming, hester2018deep, rajeswaran2017learning}, simply adding demonstrations to the replay buffer \citep{vecerik2017leveraging}, and introducing reward shaping terms which encourage RL agent to follow demonstrations \citep{tang2024automate, reddy2019sqil, koprulu2024dense, peng2018deepmimic}. Other works have sought to squeeze more information out demonstrations by automatically translating existing demonstrations to new initial conditions and scenes \citep{mandlekar2023mimicgen} or by robustifying BC policies by promoting recovery behavior \citep{ke2023ccil, ankile2024juicer}. While our work complementarily pushes the limits of what behaviors can be learned entirely from scratch, we expect that demonstrations (when available) can also be incorporated into our framework to further accelerate learning.

\section{Generating Diverse Resets for Learning Dexterous Manipulation}
\label{sec:method}

\begin{figure}[t!]
\centering
\includegraphics[width=\linewidth]{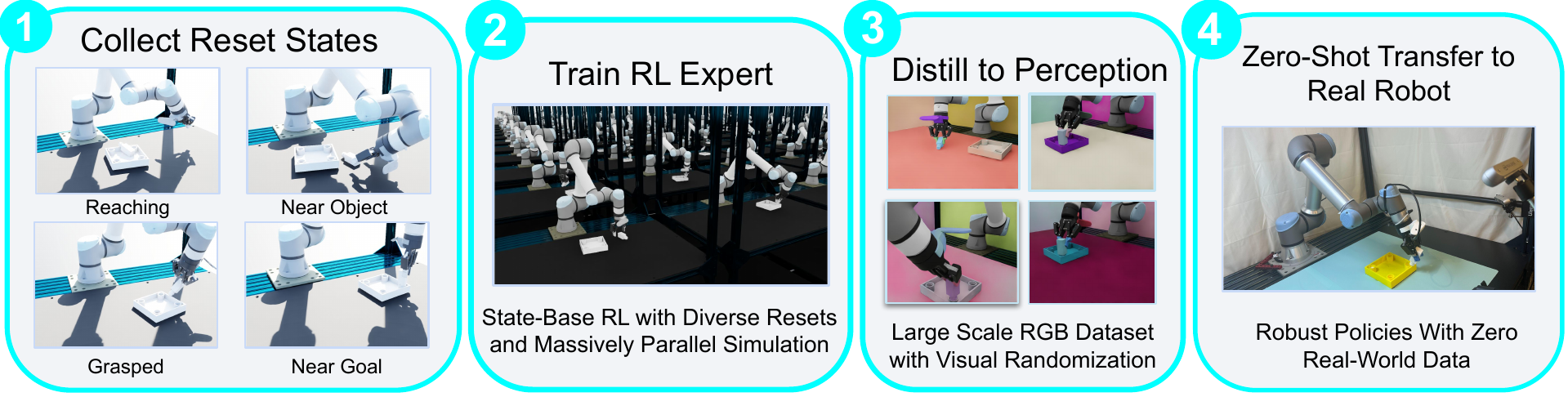}
\caption{\footnotesize{\textbf{Sim-to-Real Pipeline with \Method} (1) After generating partial assemblies and grasps from the simulator, (2) we collect reset states: reaching, near object, grasped, and near goal. (3) We then train a state-based RL policy initialized from these reset states, which is used to (4) train student-teacher distillation to get a RGB policy. (5) We finally deploy the RGB-based policy in the real world zero-shot.}}
\label{fig:methods}
\end{figure}

This section introduces \Method, a scalable framework for robotic manipulation that automatically constructs RL problems by generating diverse, manipulation-centric reset distributions in a task-agnostic way. Rather than relying on task-specific curricula, demonstrations, or carefully tuned rewards to guide exploration, \Method\ exposes standard RL algorithms to a broad superset of key interaction states that would be rarely encountered under naïve exploration. When paired with large-scale simulation and compute, this reset design continually exposes the algorithm to diverse interaction states, preventing convergence to narrow, suboptimal behaviors and enabling complex multi-step manipulation skills to emerge from large-scale optimization.  

\subsection{Problem Setting}
\textbf{Reinforcement Learning Problem:} We formalize the RL problems synthesized by \Method \ as a Markov decision process (MDP) defined by the tuple $(\mathcal{S}, \mathcal{A}, P, r, \gamma, \rho)$, where $s \in \mathcal{S}$ denotes the state, $a \in \mathcal{A}$ denotes the robot's action, $s' \sim P(\cdot | s, a)$ denotes the next state sampled from the transition dynamics, $r$ is the reward function, $\gamma$ is the discount factor, and $s_0 \sim \rho$ is the initial state distribution. We optimize for the discounted sum of rewards: $J(\pi) = \mathbb{E}_{s_0 \sim \rho, a \sim \pi}\!\left[ \sum_{t=0}^\infty \gamma^t r(s_t,a_t) \right]$, where $a \sim \pi(\cdot | s)$ denotes actions sampled from the policy. We focus our RL training on compact state representations (i.e., Lagrangian states), only moving to vision-based distillation for transfer to the real world (Sec.~\ref{sec:distillation}).

\textbf{Task Scope and User Inputs:} To effectively leverage task structure when automatically designing RL problems we make several practical assumptions. First, we focus on manipulating rigid bodies where the goal is to move a single user-specified object to a target configuration (potentially relative to other objects). For example, the assembly task in Figure \ref{fig:leg_sim_real} requires picking up a table leg, moving it to the desired hole, and screwing the pieces together---a long-horizon task requiring diverse behaviors, yet framed as moving one object to a goal. Specifically, \Method \ requires the following input from the user: 
\begin{requirement}\label{req:target}
A target object $s^{\text{tar}} \subset s$ to be manipulated.
\end{requirement}
\begin{requirement}\label{req:goal}
A set of goal configurations $\mathcal{G} \subset \mathcal{S}$ for $s^{\text{tar}}$.
\end{requirement}
\begin{requirement}\label{req:bounding}
A workspace $\mathcal{W} \subset \mathcal{S}$ for the robot.
\end{requirement}
At a code level, this requires that the user identify $s^{tar}$ in the environment definition, provide a means for sampling goal states, and select and operating region for the robot (such as the area over a table top). These pieces of information place minimal burden on the user, yet nonetheless contain rich information about problem structure which \Method \ exploits to design easily solvable RL problems. In short, \Method \ automatically constructs a diverse initial state distribution $s_0 \sim \rho$ that enables RL algorithms to solve all considered tasks using a single task-agnostic reward function, without any task-specific hyperparameter tuning or explicit curricula.

\subsection{Automatically Generating RL Problems with Diverse Resets}
\label{sec:diverse_reset}

Reinforcement learning fails to solve long-horizon manipulation tasks when training only encounters a narrow slice of potential object configurations or robot-object interactions. To prevent this collapse, \Method \ systematically expands coverage along two axes. First, we approximately cover the space of pathways along which the target object can be transported to the goal by densely resetting $s^{tar}$ on the tabletop, at random point in the air, and at and near the goal $\mathcal{G}$. This exposes the RL algorithm to difficult-to-discover success signals, and allows these signals to smoothly propagate throughout the state-space as the value function is updated. Next, we cover the different ways the robot can interact with $s^{tar}$ by resetting the arm in configurations where it is reaching towards $s^{tar}$, making contact with $s^{tar}$ from a wide variety of points, and with stable grasps on $s^{tar}$. This exposes the RL algorithm to the different ways the robot can interact with $s^{tar}$ to move it towards the goal. Altogether, this broad coverage enables the RL algorithm to discover high-value regions of the state space and the behaviors required to reach those states. In the parlance of motion planning, we are exposing the agent to states in ``narrow passages", allowing them to be easily traversed through RL-based optimization.

Specifically, we generate these resets as follows. First, we use the grasp sampler from \cite{mittal2023orbit} to calculate a set of $1000$ feasible grasp points on the target object $s^{tar}$, as depicted in Figure \ref{fig:grasps}. Next, we generate a set of feasible offsets for $s^{tar}$ relative to the  goal $\mathcal{G}$. This is accomplished by spawning $s^{tar}$ at $\mathcal{G}$, and then applying small random forces to dislodge the target from the goal, as in \cite{tang2023industreal}. For example, for the insertion task in Figure \ref{fig:leg_sim_real}, this process generates a continuum of relative configurations between the peg and the hole where the peg is only partially inserted. With these pre-computed quantities, we generate the following resets over the space of robot-object configurations: 

\begin{wrapfigure}{r}{0.6\linewidth}
    \centering
    \vspace{-0.5cm}
    \begin{subfigure}{\linewidth}
        \includegraphics[width=\linewidth]{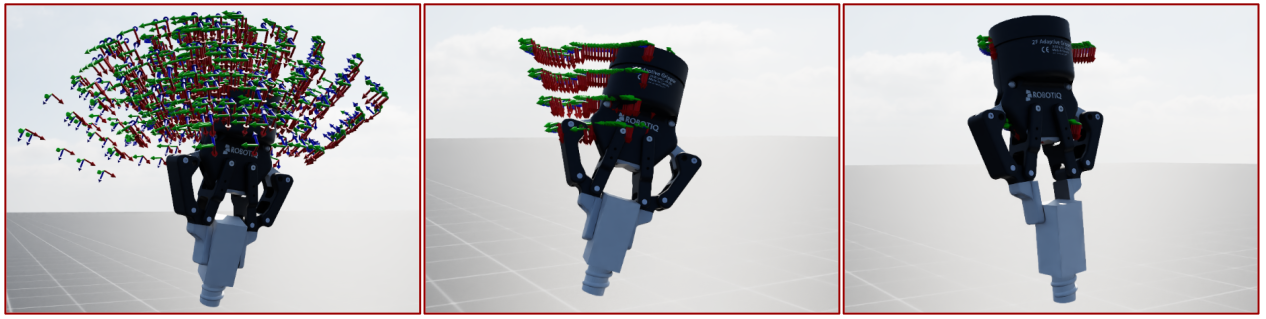}
        \caption{\footnotesize
        \textbf{Grasp Sampling.} We display the grasp poses ($S^{G} \subset S$) sampled for the table leg. From left-to-right, the grasp sampling ranges are broad, moderate, and narrow. Our method uses the broad range.}
        \label{fig:grasps}
    \end{subfigure} 
    \vspace{-1cm}
\end{wrapfigure}

\textbf{1) Reaching Resets:} $S^{R} \subset \mathcal{S}$ capture states where the robot is moving towards the target object. For our tabletop manipulation problems, this corresponds to resetting the target object at diverse poses on the table top, with the gripper spawned at random poses throughout the workspace $\mathcal{W}$.

\textbf{2) Near-Object Resets:} $S^{NO}$ also spawns $s^{tar}$ across the tabletop, but then resets the end effector to one of the pre-computed grasp points with a small random offset and randomly set the gripper to either be open or closed. This provides broad coverage over states where the robot is initiating contact with $s^{tar}$ from a wide distribution of directions, providing coverage over both non-prehensile interactions and the initiation of stable grasps.

\textbf{3) Stable Grasp Resets:} $S^{G} \subset S$ covers states where prehensile manipulation occurs and the robot has secured a stable grasp on $s^{tar}$. We spawn $s^{tar}$ randomly in the air throughout $\mathcal{W}$, then spawn the gripper at a feasible grasp point. 

\textbf{4) Near-Goal Resets:} $S^{NG} \subset S$ provides dense coverage over near goal states where contact rich behavior such as insertion or twisting occur. We spawn $s^{tar}$ at one of the pre-computed offsets from the goal $G$, then spawn the gripper to be in contact with $s^{tar}$ as with the Near-Object resets. 


Altogether, these different reset regions provide dense, approximate coverage over the space of pathways to the goal, without prescribing \emph{a priori} which behaviors are needed to solve the task. It is important to note two important things about this reset distribution- 1) we do not order or connect the states in any graph structure, the paths between them are completely determined by RL, 2) we do not inform any dynamic behavior between these states, these are completely emergent from the RL process. Indeed, as we see from our wide range of examples (Fig.\ref{fig:leg_sim_real}), the RL algorithm is free to select completely different pathways through the state space when solving different tasks, utilizing the reset states that are useful while ignoring those that are not. For example, in the \texttt{Drawer Insertion} task, at convergence, the RL algorithm does not obtain stable grasps on the drawer, but instead repeatedly flips the drawer and then pushes it into the cabinet.  In contrast, for the \texttt{Leg Twisting} tasks, the robot picks up the table leg, pushes it against the table to obtain a more favorable grasp, and then twists the leg into the hole. 

\textbf{Practical Implementation:} In practice, it can be difficult to efficiently sample \emph{feasible resets} that respect the contact constraints of the physics simulator. Sampling invalid initial conditions can lead to pathological and non-physical behavior that destabilizes learning. Moreover, reset states must be sampled with minimal overhead to maximize GPU-parallelism. Thus, we first sample feasible resets during an offline phase, sampling proposed resets from the four regions defined above and rejecting invalid samples using a combination of collision checking and stepping the simulator for a few steps to allow for stabilization. This yields four validated datasets corresponding to the four reset distributions described above: $D^{R}$, $D^{NO}$, $D^{G}$, and $D^{NG}$. During RL training, we sample from $\mathrm{Uniform}(D)$ where $D = D^C \cup D^{NO} \cup D^{G} \cup D^{NG}$. We cache these resets on-GPU to ensure efficient sampling during training. 

\subsection{Algorithmic Decisions for RL Training}
\label{sec:rl_design}
Next, we discuss key algorithmic decisions that are required for an on-policy algorithm (PPO~\cite{schulman2017proximal}) to learn manipulation behavior by leveraging the diversity of reset states and scale to the complexity of tasks we consider in this work. 

\textbf{Task-Agnostic Reward Structure:} Leveraging the design choices described above, we use a simple, common reward function shared across all tasks:
\begin{equation}
\label{eq:reward_main}
r(s_t, a_t)
=
r_{\text{success}}(s_t)
+
r_{\text{dist}}(s_t)
+
r_{\text{reach}}(s_t)
+
r_{\text{smooth}}(s_t, a_t)
+
r_{\text{term}}(s_t).
\end{equation}
Here, $r_{\text{success}}$ is a sparse binary reward indicating task completion, $r_{\text{dist}}$ encourages minimizing the distance of the target object $s^{tar}$ to the goal, $r_{\text{reach}}$ encourages the gripper to be near $s^{tar}$, $r_{\text{smooth}}$ penalizes large or rapidly changing actions, and $r_{\text{term}}$ penalizes unsafe or physically invalid states that trigger terminations. Importantly, this reward does not encode task-specific strategies as all components and weights are kept fixed across experiments. We find that this generic structure is sufficient for stable training across diverse manipulation tasks, and performance is largely insensitive to the precise weighting of individual terms. See Appendix~\ref{sec:reward} for additional details.

\textbf{Scaling Parallel Environments:} When combined with increasing reset diversity, we found that scaling the number of parallel environments used by PPO stabilized and accelerated learning. This enables \Method \ to obviate curricula and task dependent rewards, significantly reducing the amount of tuning which is required to solve a new task. Intuitively, this large batch size prevents catastrophic forgetting in situations when many of the reset states result in unsuccessful policy behavior. This is necessary to ensure continued value propagation backwards from the target configuration. 

\textbf{Asymmetric Actor-Critic:} Since we have access to privileged information in a simulator, we use an asymmetric actor-critic approach \citep{pinto2017asymmetric} for our learning architecture. The actor observations include a history of the five previous time-steps for the state of the robot, the poses of all objects in the scene, and the previous actions taken by the policy. The critic takes in these observations as well as additional privileged parameters of the environment. We found that conditioning the actor on larger observation spaces led to less stable training and led us to only provide this information to the critic. 

\textbf{Generalized State-Dependent Exploration Noise:} We employ the policy noise parameterization from gSDE \citep{raffin2022smooth}. gSDE has a separate prediction head which determines the gaussian exploration noise at each time-step and is conditioned on the features of the final layer of the policy network. This approach enables the actor to learn different temporally-correlated exploration strategies in different regions of the state-space, crucial for solving heterogeneous multi-stage tasks. 

\section{Simulation Experiments}\label{sec:experiments}

Let us first consider a study of data generation in simulation with \Method. We aim to address the following questions experimentally - (Q1) Does \Method \ outperform baselines in terms of asymptotic performance and sample complexity, enabling tasks beyond current methods? (Q2) How do the key design decisions laid out in Section~\ref{sec:rl_design} affect performance? (Q3) Can the learned RL policies be used to generate diverse data for sim-to-real transfer?

\subsection{Task Descriptions}

\begin{figure}[H]
\centering
\includegraphics[width=\textwidth]{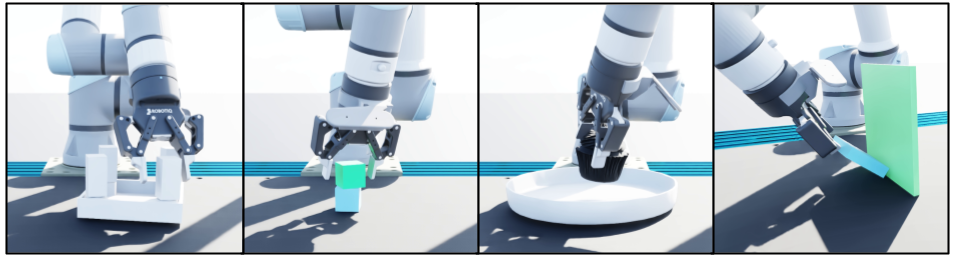}
\caption{\footnotesize
\textbf{Additional tasks.} Visualizations of the new manipulation tasks solved with OmniReset. From left to right: Four-Leg Table Assembly, Cube Stacking, Cupcake on Plate, Block Reorientation on Wall}
\label{fig:additional_tasks}
\end{figure}

We use the following tasks to demonstrate the effectiveness of the \Method \ framework. The \texttt{Leg Twisting} tasks is a replication of the the \texttt{square\_table} task from \citep{heo2023furniturebench}, and involves screwing in a single table leg. The \texttt{Drawer Insertion} task is based on the task by the same name from \citep{heo2023furniturebench} and requires inserting a drawer into a dresser. The \texttt{Peg Insertion} task requires inserting a peg into a hole. The $\texttt{Cube Stacking}$ task requires stacking one cube on top of another with a desired orientation. The \texttt{Wall Slide} task requires non-prehensile motion to push a block up against a wall into a desired orientation. Finally, the \texttt{Cupcake Placement} task requires placing a cupcake on a plate at a desired orientation. For each task, we consider both \texttt{Hard} and \texttt{Easy} variants. For the \texttt{Hard} variants, the target object $s^{tar}$ is distributed on the table with $x-y$ coordinates in $(x,y) \in [-0.2, 0.2] \times [-0.15, 0,15]$ and task-specific randomization over the goal position, whereas the \texttt{Easy} version of each task uses a highly restricted set of initial conditions with $(x,y) \in [0.1,0.12] \times [0.1, 0.12]$ with a single fixed goal location. We refer the reader to our public code release for additional task specific parameters, but summarize each task below. Snapshots from the tasks are depicted in Figures \ref{fig:leg_sim_real} and \ref{fig:additional_tasks}

Finally, to fully demonstrate the utility of \Method, we solve the \texttt{Four Leg} task depicted in Figure \ref{fig:additional_tasks} by $a)$ training independent \Method \ policies to screw in each of the four legs then $b)$ using a simple scripting policy to switch between the policies to complete the overall long-horizon task. The demonstrations highlights how \Method \ can be combined with high-level planning to push the horizon of tasks that can be solved to even greater extremes. 

\subsection{Baseline Comparisons in Simulation}

\begin{figure}[!h]
    \centering
    \includegraphics[width=\linewidth]{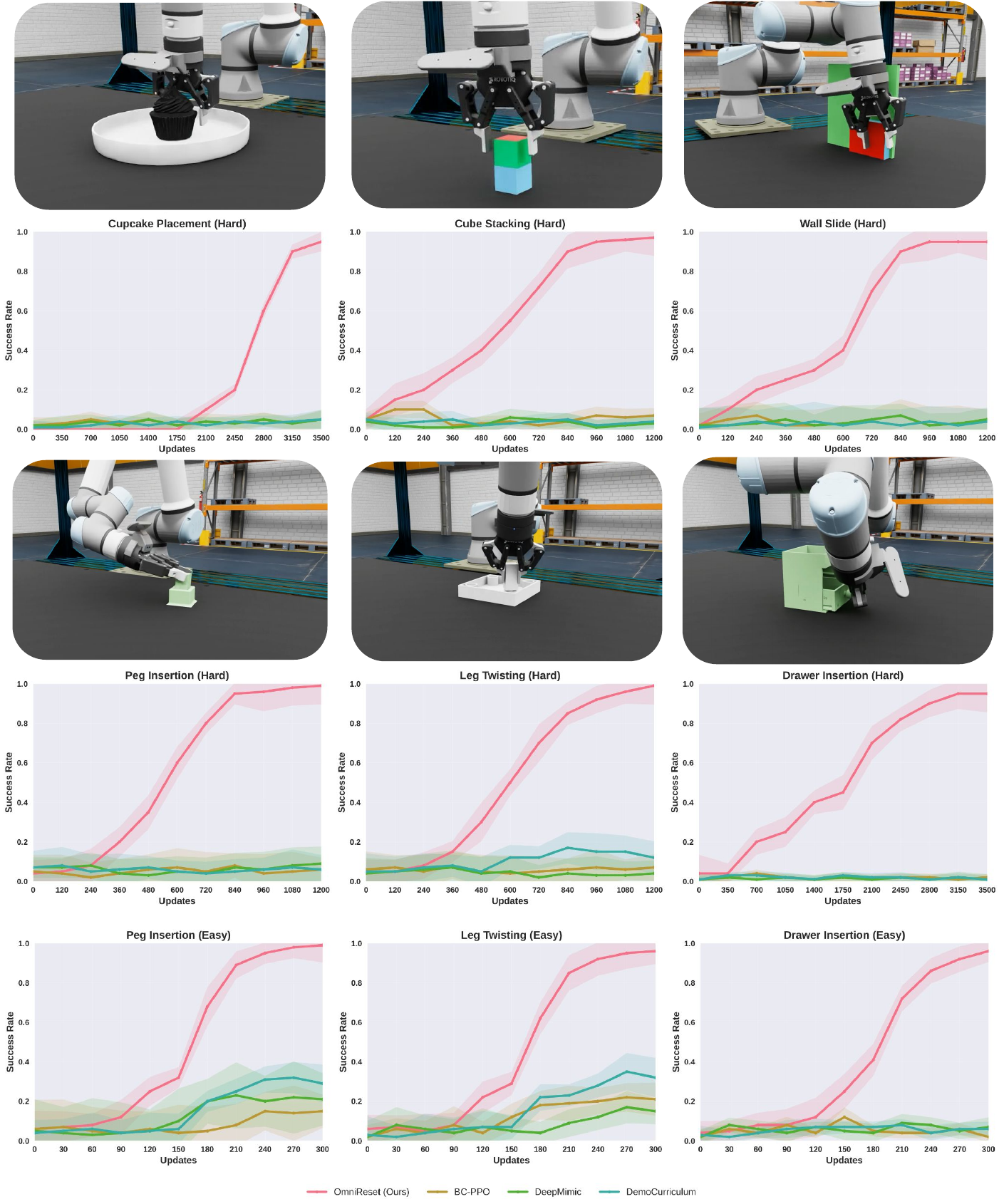}
    \caption{\footnotesize{
    \textbf{Success rates during RL training.} We plot success rates over learning process for tasks described in Sec.\ref{sec:experiments}. We see that \Method scales to task where baselines struggle to make meaningful progress, especially with the wide range of initial conditions specified in the \text{Hard} variants of tasks.}}
    \label{fig:baselines}
\end{figure}

We compare to the following baselines, which bootstrap learning with expert demonstrations, providing them with more prior information about how to solve the task than \Method. For each of the methods, we supply 100 successful demonstrations, effectively giving the baselines additional access to the optimal behavior, while \Method \ does not have this information. The initial conditions for these demonstrations are drawn from the reaching region $S^R$, which corresponds to the set of initial conditions the robot will encounter when solving the full task. 

\begin{hangingpar}
\textbf{1) BC-PPO:} We add a Behavior-Cloning (BC) loss to the PPO objective to construct a baseline emblematic of numerous works combining BC and RL objectives \citep{hester2018deep, rajeswaran2017learning}. When training this algorithm, the environment is always reset from the reaching region $S^R$, which reflects the `standard' reset distribution typically used to solve such tasks.
\end{hangingpar}

\begin{hangingpar}
\textbf{2) DeepMimic:} To compare with methods for motion imitation, we use DeepMimic-style reward augmentation \citep{peng2018deepmimic} on top of our generic rewards. During resets, a random demonstration is chosen, and the agent is reset from a random point along the demonstration and is given an auxiliary reward which provides bonuses for following the demonstration.  
\end{hangingpar}

\begin{hangingpar}
\textbf{3) Demo Curriculum:} This baseline is constructed in the spirit of method such as \citep{bauza2025demostart} and uses a success-weighted autocurriculum to sample reset states from the demonstrations. We use PPO~\citep{schulman2017proximal} as the base RL algorithm to ensure a fair comparison. 
\end{hangingpar}

\textbf{Learning Curves and Success Rates:} We plot learning curves for the \texttt{Hard} variants of each task and the $\texttt{Easy}$ variants of the \texttt{Peg Insertion}, \texttt{Leg Twisting}, and \texttt{Drawer Insertion} tasks in Figure \ref{fig:baselines}. Success rates are reported for initial conditions sampled from the Reaching resets $S^R$, which correspond to states from which the robot must solve the full task. We see that \Method \ is able to consistently obtain high success rates on each of the tasks, substantially outperforming baseline methods. While the baselines make some progress on the \texttt{Easy} variants of the tasks, they consistently struggle to scale to the wider distributions of initial conditions that define the \texttt{Hard} variants. Appendix Figure~\ref{fig:leg_appendix} provides additional fine-grained insight into the failure modes of the baselines. Here we plot success rates during training for initial conditions drawn from the demonstrations that fall into the Near-Goal and Reaching regions of the state-space. This plot demonstrates how the baselines are able to solve part of the task (i.e, when starting near the goal) but fail to scale to the full long-horizon task (i.e, starting from a reaching position). We refer readers to the supplementary website for a more detailed look at failure modes.

\begin{figure}[!h]
    \vspace{-1em}
    \centering
    \includegraphics[width=\textwidth]{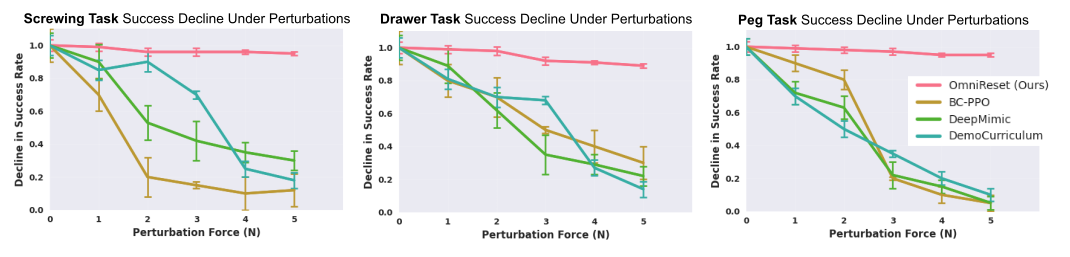}
    \caption{\footnotesize
    \textbf{Success rate over perturbations.} We plot the decline in success rate (measured by ratio of success rate between no perturbations and current level of perturbations). We find that \Method is robust to perturbations while performance of baselines drops significantly.
    }
    \label{fig:perturbations}
\end{figure}

\textbf{Emergent Curricula:} \Method \ does not rely on curricula to stabilize and accelerate learning.  We visualize how learning for \Method \ progresses over time on our \href{https://omnireset.github.io/#learning-over-time}{website}. We observe that the diverse resets \Method \ provides enables PPO to naturally solve the problem backwards, first learning to succeed from near-goal states and eventually learning to succeed from the entire search space. This behavior is entirely emergent from our diverse resets and large number of parallel environments.

\textbf{Robustness of Policies:} We conduct a robustness analysis on the learned policies (Fig~\ref{fig:perturbations}). We sample an initial condition from one of the demonstrations, perturb the initial condition with forces of different magnitudes and report policy success from these perturbed initial conditions. We find that baseline performance quickly degrades under small perturbations, while \Method \ is barely affected under large perturbations.

\begin{figure}[!h]
    \centering
    \includegraphics[width=\textwidth]{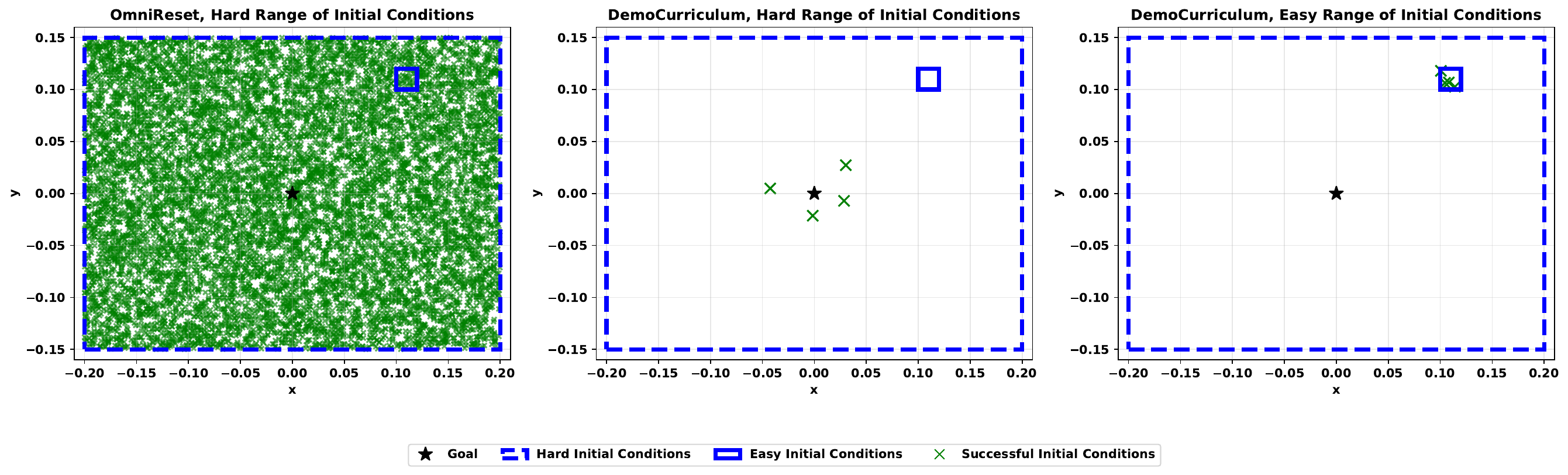}
    \caption{\footnotesize
    \textbf{Successful initial states of RL policy.} For the \texttt{Lew Twisting}, we plot the xy configurations from which RL polices trained with Demo Curriculum and \Method succeed. We find that \Method succeeds from a much broader range of initial conditions.}
    \label{fig:baselines_scatter}
\end{figure}

We also analyze the ability of policies to solve the task from a wide range of initial conditions in the scatter plots in Figure ~\ref{fig:baselines_scatter}. These plots show success rates over various initial conditions for both \Method and Demo Curriculum (the most successful baseline) on the \texttt{Leg Twisting Easy} task. For each plot we show the success rate over 1000 sampled initial conditions from the full task distribution. This plot demonstrates how the baseline struggles to achieve achieve consistent success across the distribution of initial conditions it was trained on, while \Method \ achieves is able to succeed across the entire workspace.

\subsection{Ablating Key Design Decisions }

\begin{figure}[!h]
    \centering
    \includegraphics[width=0.9\textwidth]{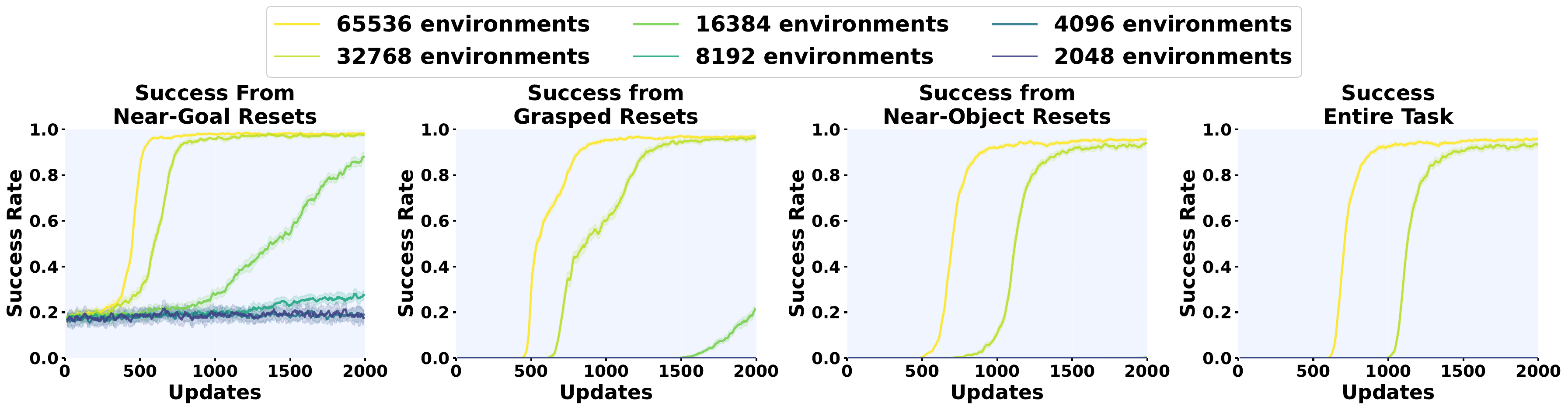}
    \caption{\footnotesize
    \textbf{Ablation on number of environments.} We plot the success rates over course of RL using different number of environments. We find that the number of environments significantly impacts training performance.}
    \label{fig:num_env_plots}
\end{figure}

We ablate $1)$ the number of parallel environments (and PPO batch size) and $2)$ the range of reset randomization used by \Method on the \texttt{Leg Twisting Hard} task. Figure \ref{fig:num_env_plots} ablates the number of parallel environments and shows success rates during training from the four different reset distributions used by \Method. The reset distributions in Figure \ref{fig:num_env_plots} are roughly ordered from left to right from the end of the task (Near-Goal) to the beginning of the task (Reaching). While runs with a smaller number of environments can make progress at solving the task from Near-Goal states, we observe that a large number of parallel environments are essential for scaling to the complexity of the full multi-stage task (i.e. from reaching states). Similarly, we see that increased diversity in the grasps used by \Method \ has a substantial effect on sample efficiency, highlighting how densely covering the different modes of robot-object interactions is crucial for efficient and scalable RL training.

\begin{wrapfigure}{r}{0.5\textwidth}
    \centering
    \vspace{-2.0cm}
    \includegraphics[width=\linewidth]{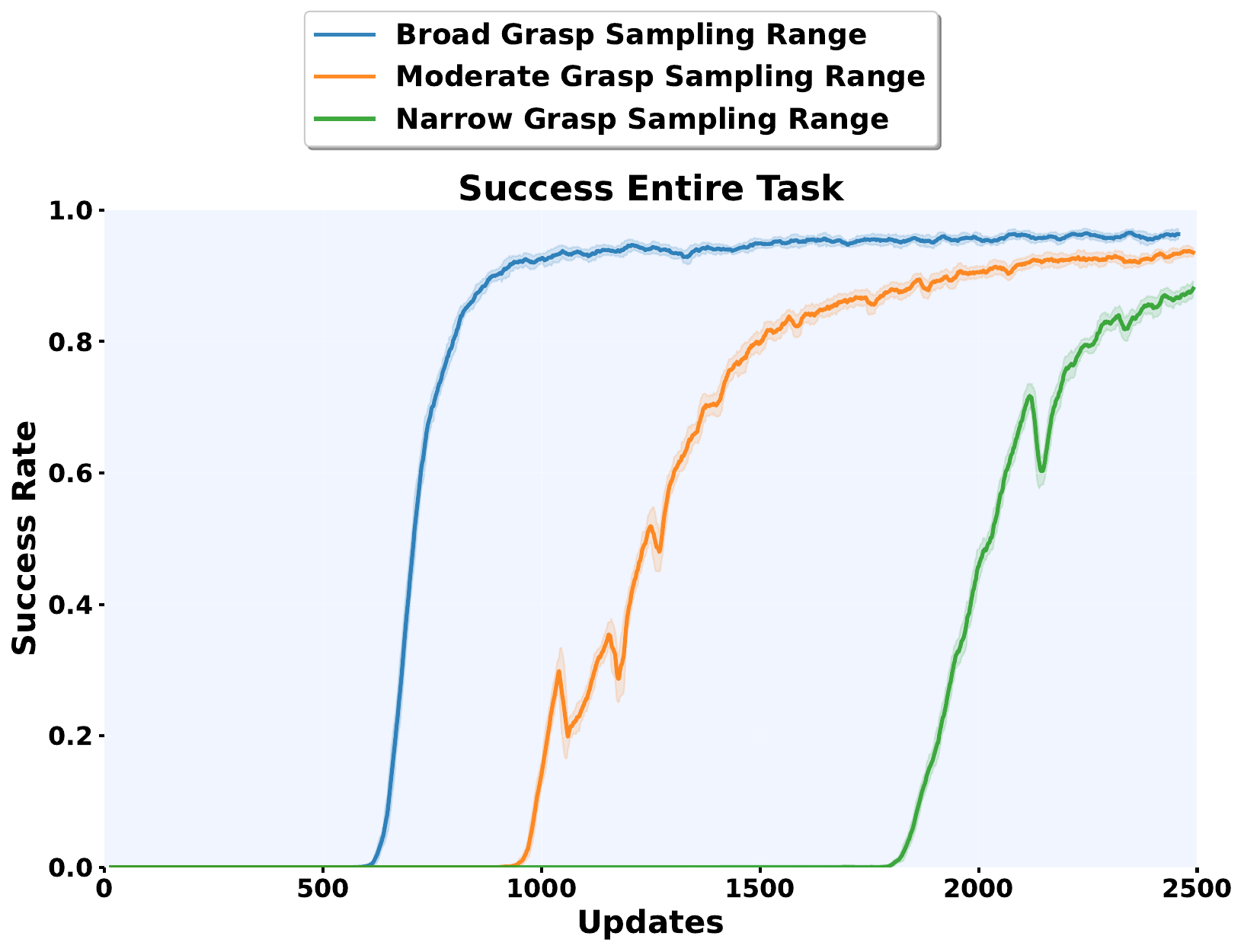}
    \caption{\footnotesize
    \textbf{Ablation on grasp sampling range.}
    For this ablation on the \texttt{Leg Twisting} task, we find that training RL on narrower grasp sampling ranges leads to worse sample efficiency and lower success rate.}
    \label{fig:grasp_sampling_ablation}
    \vspace{-0.4cm}
\end{wrapfigure}

\section{Distillation and Real-World Transfer}

\vspace{-.5em}

\label{sec:distillation}
We demonstrate the utility of our learned data-generation policies by distilling them into visuomotor policies deployable directly on hardware from RGB inputs. Experiments are conducted on a UR7e robot equipped with a Robotiq 2F-85 gripper, with control and policy inference running on a PC with an RTX 4090 GPU. The robot observes $224 \times 224$ RGB images from three RealSense cameras: a D455 providing the front view, a D435 for the side view, and a D415 mounted on the wrist. Using the photorealistic rendering capabilities of IsaacLab \citep{mittal2023orbit}, we collect 80,000 expert rollouts with synchronized images and actions for standard student–teacher distillation~\cite{chen21inhand}. The student policy operates at 10Hz and uses an ImageNet-pretrained ResNet-18 encoder and a Gaussian MLP head conditioned on the five most recent observations. See Appendix~\ref{sec:appendix_transfer} for full implementation details for distillation and transfer.

\begin{wrapfigure}{r}{0.5\textwidth}
    \centering
    \includegraphics[width=0.5\textwidth]{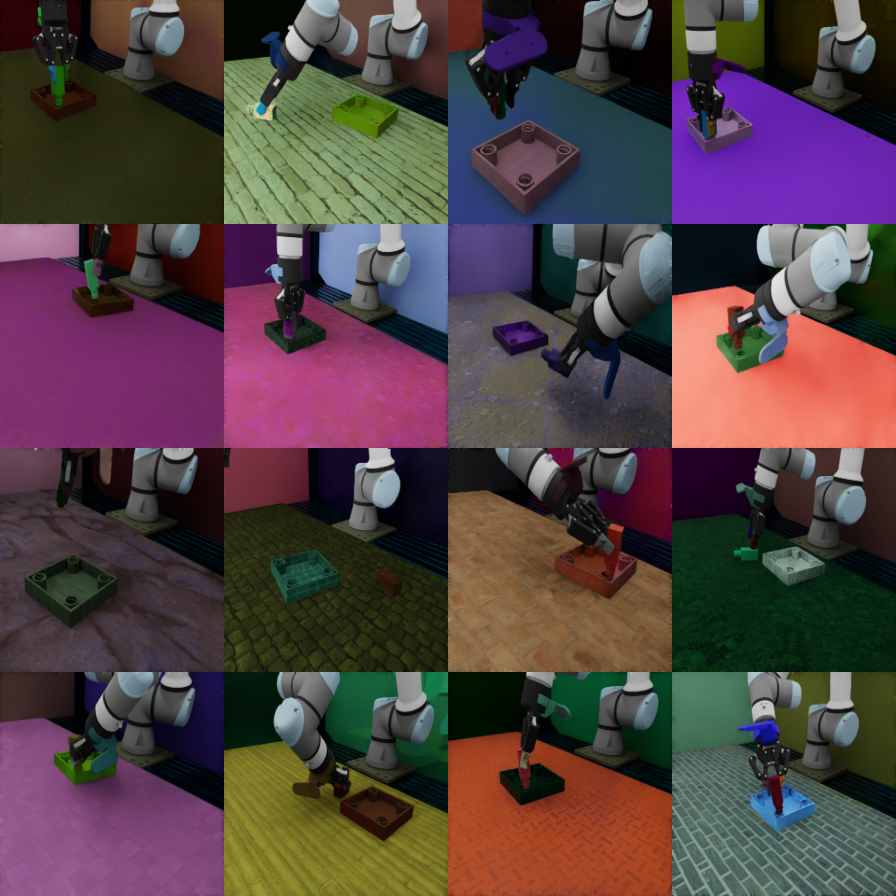}
    \caption{\footnotesize
    \textbf{Visual randomizations.}
    Examples of domain randomization applied during training. We vary lighting conditions, backgrounds, object and robot appearance, and workspace textures, along with camera pose and field-of-view jitter, to improve robustness to real-world visual variation.}
    \label{fig:visual_rand}
    \vspace{-0.4cm}
\end{wrapfigure}

\textbf{Visual randomization.} As shown in Fig.~\ref{fig:visual_rand}, to mitigate the sim-to-real visual gap, we apply extensive domain randomization following DextrAH-G \citep{singh2025dextrahrgbvisuomotorpoliciesgrasp}, varying lighting, backgrounds, object and robot appearance, and workspace textures. Camera extrinsics are calibrated to the real setup with additional pose and FOV jitter for robustness. We also apply standard image augmentations including color jitter, blur, grayscale, and noise.

\begin{wrapfigure}{r}{0.5\textwidth}
    \centering
    \includegraphics[width=0.5\textwidth]{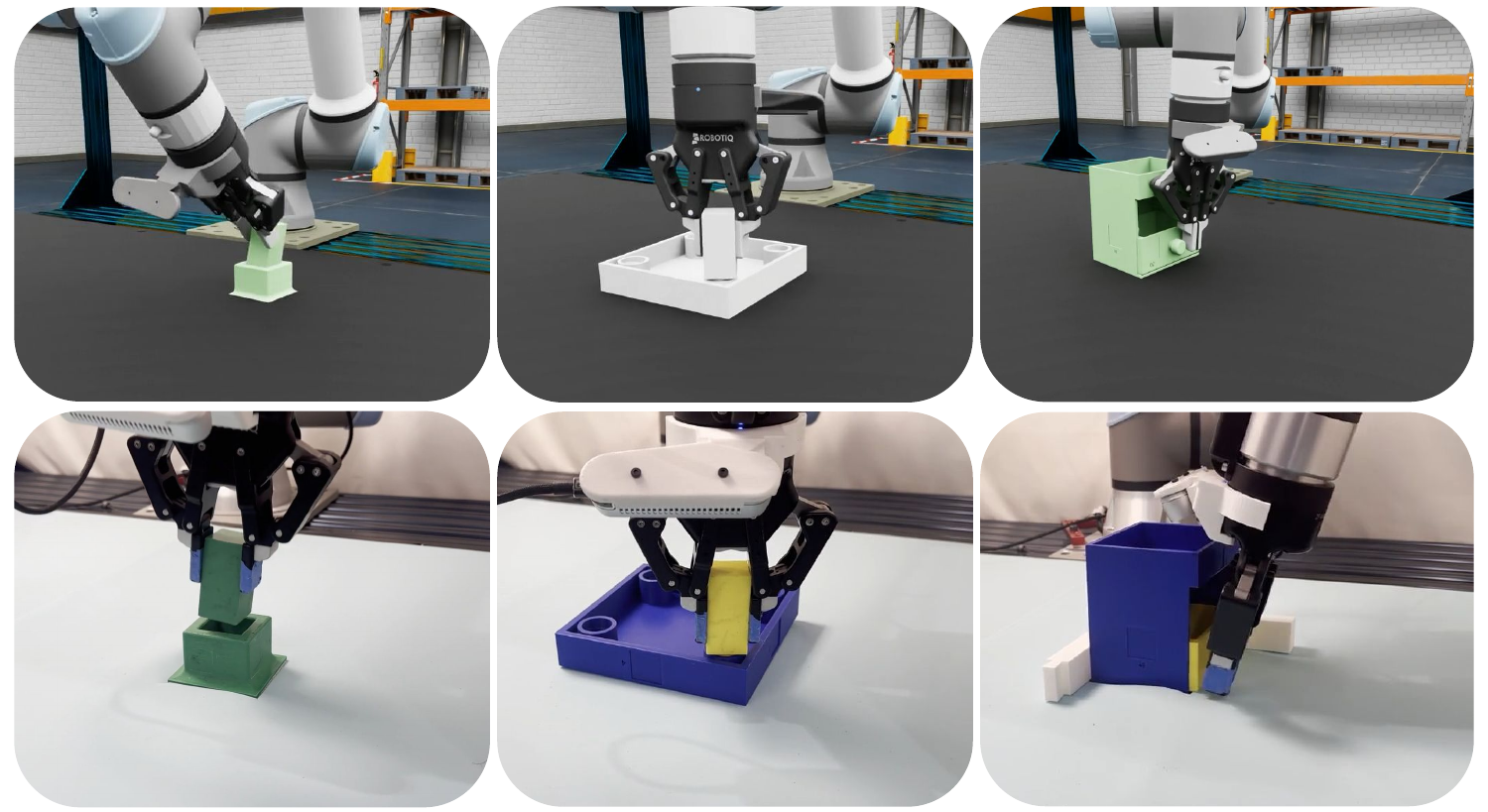}
    \caption{\footnotesize
    \textbf{Sim2Real Tasks.}
    Simulation setup for pretraining (top) and real-world deployment (bottom) of peg insertion (left), leg twisting (middle), and drawer assembly (right).}
    \label{fig:sim2real_3tasks}
    \vspace{-0.4cm}
\end{wrapfigure}

\textbf{Dynamics randomization.} To reduce the control gap, we calibrate kinematics to hardware and deploy the same task-space operational space controller \citep{Khatib1987OperationalSpace} in simulation and reality, with the policy predicting end-effector pose deltas that are converted to torques using identical Jacobian-based control. We perform system identification of key actuator parameters (friction, armature, and delay) to match real joint behavior following PACE \citep{bjelonic2025towards}, then randomize controller gains and physical parameters around identified values during RL training. We additionally randomize object mass and friction to improve robustness to contact dynamics, and apply a curriculum that progressively reduces the action space to encourage smaller, smoother motions. We emphasize that this curricula was used to improve sim-to-real transfer, and is not necessary for learning high-performing policies in simulation. Indeed, the simulation only evaluations we report are trained without the curricula, maintaining our claim that \Method \ does not rely on complex training curricula for behavior generation.

\textbf{Real-World Transfer.} We successfully deploy a distilled \Method\ policy on the \texttt{Peg Insertion}, \texttt{Leg Twisting}, and \texttt{Drawer Insertion} tasks, as shown in Figure~\ref{fig:sim2real_3tasks}. The distilled RGB policies, trained on 80{,}000 simulation trajectories spanning a wide range of initial conditions, achieve zero-shot real-world success rates of \textbf{85.37\%} on Peg, \textbf{56.36\%} on Leg, and \textbf{15.38\%} on Drawer, as summarized in Table~\ref{tab:sim2real_results}. This substantially outperforms a behavior cloning Diffusion Policy baseline~\cite{chi2025diffusion}, which achieves $\sim$2\% success across tasks when trained on 100 demonstrations.

We additionally report \emph{first-try success}, defined as completing the task with a single grasp and execution (i.e., no dropping and regrasping), as well as \emph{throughput}, measured as successful completions per minute. These metrics highlight that there remains significant room for improvement in efficiency and reliability. Full real-world evaluation details are provided in Appendix~\ref{sec:appendix_evaluation}.

Qualitatively, the \Method\ visuomotor policy exhibits robust retrying behavior, recovering from initial failures and successfully completing the task. For examples of these behaviors, see \href{https://omnireset.github.io/#evaluations}{our project website}. Overall, these results demonstrate that \Method\ provides a scalable foundation for training sim-to-real policies capable of handling substantially broader initial condition distributions than prior approaches.

\begin{table}[!h]
    \caption{\footnotesize{\textbf{Sim-to-real performance across tasks:} We report state-based RL performance in simulation, distilled image policy performance in simulation and real-world deployment, and a behavior cloning baseline trained on only 100 real-world demos.}}
    \centering
    \begin{tabular}{lccc}
    \toprule
    \textbf{Metric} & \textbf{Peg} & \textbf{Leg} & \textbf{Drawer} \\
    \midrule
    State RL Success Rate (Sim) & 94.97\% & 93.94\% & 89.60\% \\
    Distilled Image Policy Success Rate (Sim) & 55.45\% & 43.69\% & 84.00\% \\
    Distilled Image Policy Success Rate (Real) & 85.37\% & 56.36\% & 15.38\% \\
    Distilled Image Policy First-Try Success Rate (Real) & 21.95\% & 43.64\% & 5.77\% \\
    Distilled Image Policy Throughput (success/min, Real) & 1.49 & 0.63 & 0.31 \\
    Real-Only BC (100 demos) Success Rate (Real) & 2.44\% & 1.82\% & 1.92\% \\
    \bottomrule
    \end{tabular}
    \label{tab:sim2real_results}
\end{table}

\section{Conclusion}
In this work we presented \Method, a simple and scalable system for data generation in simulation for complex, dexterous tasks. The primary insight in \Method \ is showing that a diverse, minimally structured set of reset states paired with large-batch on-policy reinforcement learning in simulation can lead to the emergence of surprisingly complex dexterous behavior. We provide a general purpose recipe to instantiate data generators across a variety of manipulation tasks, and demonstrate both the efficacy of this paradigm in simulation and it's ability to train robust policies which can be successfully transferred directly to the real world. However, the preset framework has several limitations which leave the door open for exciting future works. \Method \ is dependent on the quality of grasps obtained by the grasp sampler, which can fail to generate diverse grasps on complex, highly non-convex objects. Moreover, pre-computing stable grasps for tasks which require bimanual manipulation or dexterous hands will present additional challenges, and whether the techniques and principles behind \Method \ will scale to such settings remains an open question. Moreover, \Method \ uses relatively modest levels of dynamics randomization for RL training compared existing sim-to-real approaches, and we expect there to be additional challenges training policies which must adapt their behavior to solve the task over a wide range of potential operating conditions. For distillation and transfer, even in simulation our RGB policies achieve much lower success rates than the state-based experts. We expect that reaching higher success rates will require additional research into how to best combine techniques such as DAgger and RL directly from images. Moreover, we found that scaling the RGB dataset to be as large as possible continually improved performance up to the $80k$ trajectories we were able to train on with our compute budget, and we believe obtaining a deeper understanding of scaling laws for this setting will prove valuable. \cite{xu2023unidexgrasp}.

\section{Reproducibility Statement}
We have made efforts to ensure reproducibility of our results by describing the steps of our data generation and training pipeline (Sec. \ref{sec:method}), our distillation and transfer pipeline (Sec. \ref{sec:distillation}), and our experimental results (Sec. \ref{sec:experiments}). Additional ablation studies are provided in the Appendix.

\section{Acknowledgements}
We would like to thank Arhan Jain, Sriyash Poddar, Marius Memmel, and Emma Romig for their invaluable engineering help, and Mateo Guaman Castro for assistance with creative real-robot videos. We also thank Filip Bjelonic, Bingjie Tang, and Iretiayo Akinola for research discussions and advice. Finally, we thank Chuning Zhu, Prashanth Rajan, Arhan Jain, Mateo Guaman Castro, Entong Su, Marius Memmel, Jesse Zhang, David Celis Garcia, and Brenda Potts for their gracious feedback on the website design. Patrick Yin is supported by the National Science Foundation Graduate Research Fellowship (NSF GRFP). Patrick Yin and Tyler Westenbroek have also been supported by National Science Foundation under Grant No. 2212310.



\bibliography{iclr2026_conference}
\bibliographystyle{iclr2026_conference}

\newpage
\appendix

\section{Appendix}

\subsection{Detailed Reward Specification}
\label{sec:reward}

We provide the full specification of the task-agnostic reward introduced in Equation~\ref{eq:reward_main}. The reward consists of a weighted sum of safety and task-related terms that are shared across all environments, with identical weights and no task-specific tuning.

\paragraph{Safety and regularization terms.}
To encourage stable and physically plausible behavior, we penalize large actions, rapid changes in actions, and excessive joint velocities:
\begin{align}
r_{\text{smooth}}(s_t, a_t)
&=
- \lambda_{\text{act}} \|a_t\|_2^2
- \lambda_{\text{rate}} \|a_t - a_{t-1}\|_2^2
- \lambda_{\text{vel}} \|\dot{q}_t\|_2^2.
\end{align}
Additionally, we include a large penalty for invalid or unsafe robot states:
\begin{equation}
r_{\text{term}}(s_t) = -\lambda_{\text{abnormal}} \cdot \mathbf{1}[\text{abnormal state}].
\end{equation}
In practice, we use weights $\lambda_{\text{act}} = 10^{-4}$, $\lambda_{\text{rate}} = 10^{-3}$, $\lambda_{\text{vel}} = 10^{-2}$, and $\lambda_{\text{abnormal}} = 100$.

\paragraph{Task-related terms.}
We include several dense reward components that guide the agent toward successful completion of the task.

First, we encourage the end-effector to approach the target object:
\begin{equation}
r_{\text{reach}}(s_t)
=
\lambda_{\text{reach}} \left(1 - \tanh\left(\frac{\|p_{\text{ee}} - p_{\text{obj}}\|}{\sigma}\right)\right),
\end{equation}
where $p_{\text{ee}}$ and $p_{\text{obj}}$ denote the end-effector and target object positions, respectively.

Second, we provide a dense shaping signal based on the relative pose between the manipulated object and the goal:
\begin{equation}
r_{\text{dist}}(s_t)
=
\lambda_{\text{dist}} \cdot \frac{1}{2}
\left[
\exp\left(-\frac{\|x_{\text{err}}\|}{\sigma}\right)
+
\exp\left(-\frac{\|\theta_{\text{err}}\|}{\sigma}\right)
\right],
\end{equation}
where $x_{\text{err}}$ and $\theta_{\text{err}}$ denote position and orientation errors computed in the goal frame.

We also include a sparse success reward:
\begin{equation}
r_{\text{success}}(s_t)
=
\lambda_{\text{success}} \cdot \mathbf{1}[\text{position and orientation thresholds satisfied}],
\end{equation}
which activates when both position and orientation errors fall below predefined thresholds.

\paragraph{Implementation details.}
The full reward is implemented as a weighted sum of the above terms with fixed coefficients across all tasks. In particular, we use:
\[
\lambda_{\text{reach}} = 0.1,\quad
\lambda_{\text{dist}} = 0.1,\quad
\lambda_{\text{success}} = 1.0.
\]

Notably, all reward components and weights are shared across tasks without modification. We find that performance is largely insensitive to moderate changes in these weights, and that the primary driver of successful learning is the coverage of the initial state distribution rather than reward shaping.

\newpage
\subsection{Additional Simulation Results}

\begin{figure}[!h]
    \centering  
    \includegraphics[width=\linewidth]{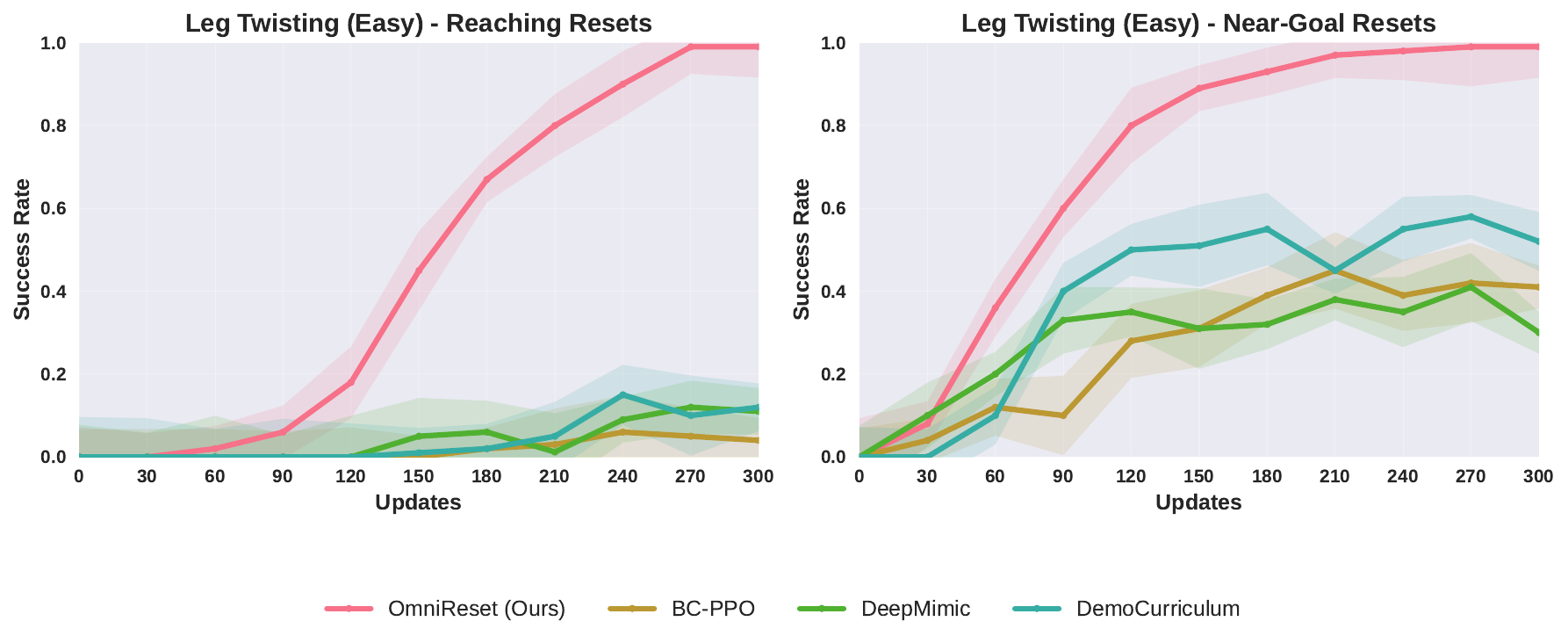}
    \caption{\footnotesize{
    \textbf{Success Rates on Different Stages of Task.} We plot the success rates for the \texttt{Leg Twisting} task when starting from states that are in the Near-Goal region and also the Reaching Region of the state space. When evaluating these success rates, we sample resets from the demonstrations used by the baseline algorithms to ensure the resulting policies start from in-distribution states.  We see that the baselines can achieve moderate success rates when starting close to the goal (Near Goal), but struggle to make meaningful progress on the full long-horizon task (captured by the reaching resets). }}
    \label{fig:leg_appendix}
\end{figure}

\subsection{Distillation and Transfer Details}
\label{sec:appendix_transfer}
\subsubsection{Robot Kinematics}
Accurate modeling of robot kinematics is a prerequisite for sim-to-real transfer. We use a Universal Robots UR7e for all real-world experiments. However, each physical 
UR7e deviates from its nominal model due to manufacturing tolerances in link lengths and joint offsets. As a result, using the default URDF leads to systematic pose errors that accumulate along the kinematic chain.

To address this, we regenerate the URDF using the factory-provided calibration file specific to our robot. This ensures that forward kinematics in simulation closely match the real system. Without this calibration step, we observe consistent end-effector pose misalignment, which significantly degrades performance on precision tasks such as insertion.

For safety, we align joint limits between simulation and the real system, with one exception for the wrist joints. Specifically, we reduce the wrist joint limits from $[-360^\circ, 360^\circ]$ to $[-180^\circ, 180^\circ]$ in simulation. This prevents the policy from exploiting extreme joint rotations during training, which could lead to the robot hitting joint limits during real-world execution and triggering safety stops.

In addition, we augment the wrist camera mount region with an enlarged, invisible collision geometry in simulation. This acts as a safety buffer to discourage behaviors that bring the wrist-mounted camera into contact with the environment. We find that these conservative modifications improve the reliability of deployment without negatively impacting task performance.

\subsubsection{Robot Dynamics and System Identification}

\begin{figure}[!htbp]
    \centering
    \begin{subfigure}{\linewidth}
        \centering
        \includegraphics[width=\linewidth]{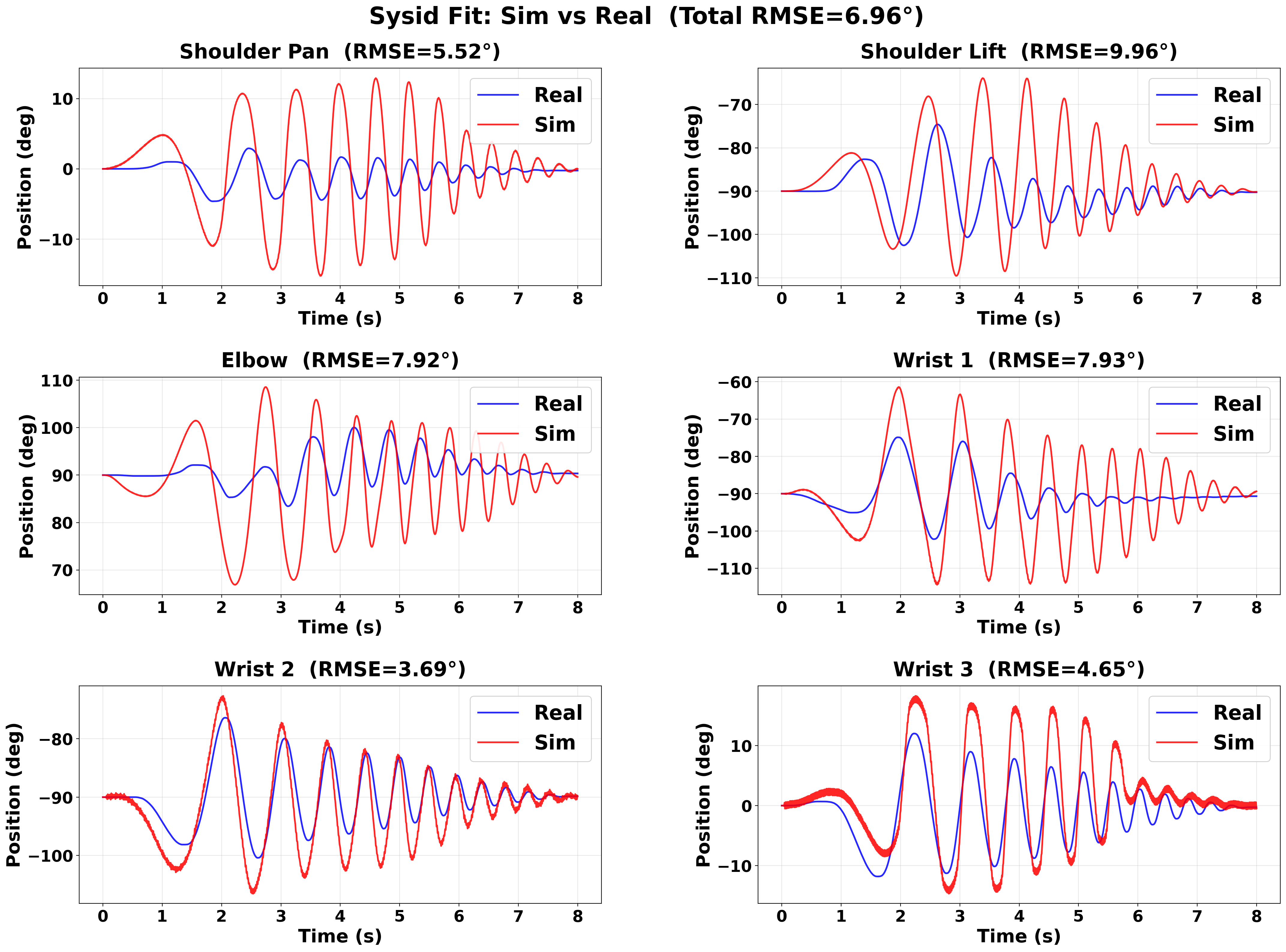}
        \caption{\footnotesize Without system identification}
        \label{fig:sysid_fit_zero}
    \end{subfigure}

    \vspace{4pt}
    
    \begin{subfigure}{\linewidth}
        \centering
        \includegraphics[width=\linewidth]{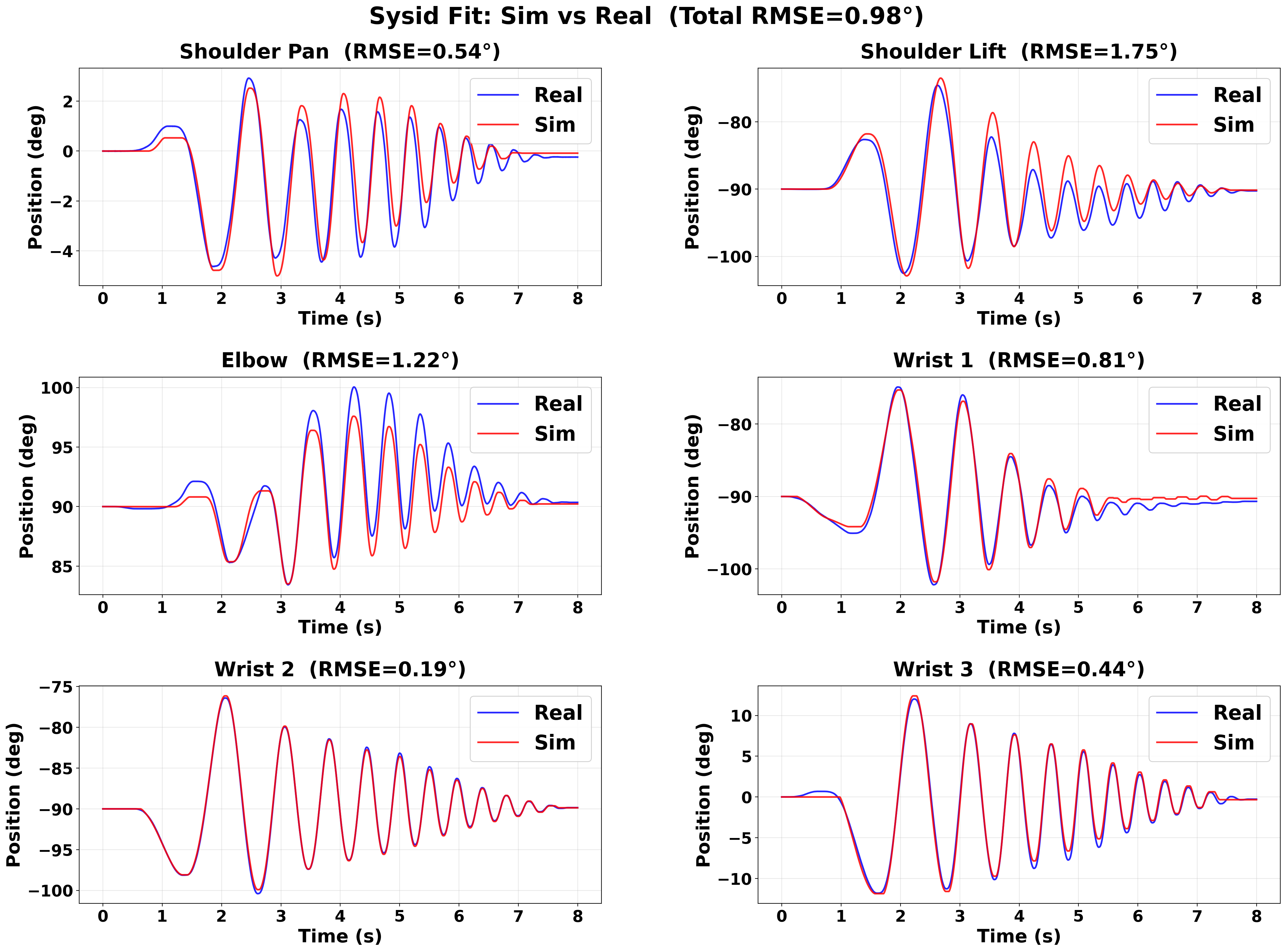}
        \caption{\footnotesize With system identification}
        \label{fig:sysid_fit}
    \end{subfigure}
    
    \caption{\footnotesize
    \textbf{Effect of system identification.}
    Comparison of chirp trajectory fit with and without system identification.}
    \label{fig:sysid_compare}
\end{figure}

Even with matched kinematics, sim-to-real transfer fails without accurate modeling of robot dynamics and actuation. A key requirement is that the interface between the policy and the robot, which maps low-frequency position outputs to high-frequency joint torques, is identical in simulation and on hardware. 

We therefore re-implement the same controller with identical gains in both simulation and real-world execution, ensuring identical computation of torques from end-effector commands. Small implementation details are critical. For example, Jacobian computations must be consistent. We found that default simulator choices, such as computing Jacobians at link centers of mass, can differ from real implementations and lead to subtle but important discrepancies. 

After aligning the controller, residual mismatch remains due to unmodeled actuator dynamics. In particular, the UR7e exhibits significant joint friction. We perform system identification of actuator parameters following PACE \citep{bjelonic2025towards} by executing open-loop trajectories and minimizing the error between simulated and real joint trajectories using CMA-ES. 

We identify friction, armature, and motor delay parameters. To ensure sufficient excitation, we use chirp trajectories that span a wide range of velocities. After system identification, we achieve an under 2 degree RMSE in joint space. In contrast, if we use zero-defaults for these parameters, we have approximately a 7 degree RMSE in joint space (see Fig.~\ref{fig:sysid_compare} for details). 

This level of accuracy is sufficient for successful transfer in tasks such as \texttt{Peg Insertion}. However, more sensitive tasks, such as flipping a drawer in the \texttt{Drawer Insertion} task, highlight that even small residual control errors can lead to significant performance degradation and failure. Reducing this control gap between simulation and hardware remains an important direction for future work.

\subsubsection{Gripper Modeling}

We use a Robotiq 2F-85 gripper mounted on the UR7e. Accurate modeling of the gripper is important for transfer, particularly for contact-rich manipulation.

Due to limitations of the physics engine, since PhysX does not support closed kinematic chains, we approximate the gripper using mimic joints in simulation. If the gripper is instead modeled with unconstrained joints, the resulting system behaves as a compliant mechanism, similar to a spring-damper system, which does not match the real hardware. In this case, the policy can exploit unrealistic deformation modes that do not transfer to the real world.

To avoid this issue, we enforce rigid coupling via mimic joints. We use a simplified binary gripper action, open or close, rather than modeling the full internal control loop. While this introduces some mismatch relative to the real system, we find that it is not performance-limiting for our tasks. In practice, careful tuning of gripper force and speed parameters is required to approximately match the effective behavior in simulation.

\subsubsection{Contact Modeling}

We use signed distance fields (SDFs) for contact modeling of objects, which is critical for accurately representing contact geometry in assembly tasks such as threaded insertions. Compared to simpler collision approximations, SDF-based contacts provide more precise surface interactions, which we find important for successful sim-to-real transfer.

We also find that the simulation timestep plays a key role in contact fidelity. Using a simulation timestep of $\texttt{sim\_dt} = 1/120$ provides a good balance between computational efficiency and accurate contact modeling for the tasks considered in this work.

Despite these improvements, and the use of extensive randomization (Section~\ref{sec:randomizations}), certain contact-rich behaviors remain difficult to transfer reliably. For example, strategies that rely on exploiting fine contact interactions, such as reorienting a peg by pushing it against the back of the hole, often fail due to small discrepancies in contact dynamics. Improving contact modeling to better capture such interactions remains an important direction for future work, including exploring more accurate contact models or integrating multiple physics simulation backends.

\subsubsection{Controller Design}

Controller design plays a central role in both policy learning and sim-to-real transfer. We use a torque-level operational space controller \citep{Khatib1987OperationalSpace} that tracks a desired end-effector pose.

Given current joint positions $q$ and velocities $\dot{q}$, we compute the end-effector pose error $e \in \mathbb{R}^6$, consisting of position and orientation error (represented as axis-angle). The desired task-space force is:

\begin{equation}
F = K_p \, e - K_d \, \dot{x},
\end{equation}

where $\dot{x} = J(q)\dot{q}$ is the end-effector velocity, $J(q)$ is the Jacobian, and $K_p, K_d$ are diagonal stiffness and damping matrices.

Joint torques are then computed via:

\begin{equation}
\tau = J(q)^\top F.
\end{equation}

We additionally clip the pose error to bound the maximum task-space force, and apply torque limits at execution time:
\begin{equation}
\tau \leftarrow \mathrm{clip}(\tau, -\tau_{\max}, \tau_{\max}).
\end{equation}

Importantly, we do not apply torque clipping during simulation training, as we found that saturation adversely affects learning performance by distorting the policy's action-to-dynamics mapping. Instead, torque limits are enforced only during real-world deployment for safety and hardware protection. 

In practice, we use separate gains for translational and rotational control. These gains are tuned empirically via teleoperation to achieve smooth, stable, and responsive behavior. For damping, we follow a critical damping heuristic and set:
\begin{equation}
K_d = 2 \sqrt{K_p},
\end{equation}
which ensures fast convergence without oscillations under a second-order system approximation. We command joint torques to the robot at 500Hz with policy end effector pose commands updating at 10Hz.

In contrast, joint-space PD control or inverse kinematics combined with joint PD results in significantly worse performance. These controllers are more prone to jamming and unstable motion. Empirically, both RL training and distilled policy performance are substantially worse under joint-space control formulations. A deeper investigation into the interaction between controller design and policy learning remains an important direction for future work.

\subsubsection{Action Space}
We use a relative Cartesian action space, where the policy predicts incremental end-effector pose updates. We find that this choice is critical for stable RL training. In particular, using small action magnitudes significantly improves exploration and learning. In contrast, absolute Cartesian actions or large action scales lead to unstable behavior, where the robot exhibits large, wild motions and fails to learn.

During training, we use an action scale of $(0.02, 0.02, 0.02, 0.02, 0.02, 0.2)$ for $(x, y, z, r_x, r_y, r_z)$. We assign a larger scale to the final rotational dimension to encourage exploration of twisting motions, which are important for tasks such as leg twisting, while remaining beneficial for other tasks. More generally, we find that larger action magnitudes are better tolerated at the wrist, where 
motions have more localized effects, whereas large motions at the base joints can lead to unstable global movements.

For sim-to-real transfer, we further reduce the action scale to 
$(0.01, 0.01, 0.002, 0.02, 0.02, 0.2)$. This reduction improves stability and mitigates aggressive motions that can degrade performance on hardware. In particular, we decrease the translational components, especially along the vertical ($z$) axis, to prevent the gripper from applying excessive forces during contact.

To avoid destabilizing the learned policy, we do not switch directly to the smaller action scale. Instead, we apply a curriculum that gradually transitions from the original training scale to the reduced scale, conditioned on task success. This allows the policy to adapt smoothly to the more conservative control regime while preserving previously learned behaviors. Note that this is not a curriculum over behaviors or state distributions, since the policy is already succeeding, but rather a parameter curriculum designed specifically to facilitate stable sim-to-real transfer.

We also note that simply clipping actions to enforce smaller motions is not effective, as it introduces discontinuities that degrade optimization and can cause RL training to collapse. Gradually adjusting the action scale through a curriculum provides a more stable alternative.

\begin{table}[t]
\centering
\small
\begin{tabular}{lll}
\toprule
\textbf{Category} & \textbf{Parameter} & \textbf{Range / Distribution} \\
\midrule

\multirow{8}{*}{Dynamics}
& Robot friction (static) & $U(0.3, 1.2)$ \\
& Robot friction (dynamic) & $U(0.2, 1.0)$ \\
& Insertive object friction & $U(1.0, 2.0)$ (static), $U(0.9, 1.9)$ (dynamic) \\
& Table friction & $U(0.3, 0.6)$ (static), $U(0.2, 0.5)$ (dynamic) \\
& Robot mass scale & $\times U(0.7, 1.3)$ \\
& Insertive object mass & $U(0.02, 0.2)$ kg \\
& Receptive object mass scale & $\times U(0.5, 1.5)$ \\
& Table mass scale & $\times U(0.5, 1.5)$ \\

\midrule
\multirow{5}{*}{Actuation / Sys-ID}
& Joint friction, armature, damping & $\times U(0.8, 1.2)$ \\
& Motor delay & $U(\{0,1,2\})$ steps \\
& OSC gains & $\times U(0.8, 1.2)$ \\
& Gripper stiffness & $\times \log U(0.5, 2.0)$ \\
& Gripper damping & $\times \log U(0.5, 2.0)$ \\

\midrule
\multirow{2}{*}{State Reset}
& Object pose 
& $x \sim U(0.3,0.55),\ y \sim U(-0.1,0.5),\ z \sim U(0,0.3),\ \mathrm{SO(3)}$ \\

& End-effector pose 
& $x \sim U(0.3,0.7),\ y \sim U(-0.4,0.4),\ z \sim U(0,0.5)$, \\
& 
& pitch $\sim U(\pi/4,3\pi/4)$, yaw $\sim U(\pi/2,3\pi/2)$ \\

\midrule
\multirow{6}{*}{Visual (Camera)}
& Third person camera position & $\pm 5$ cm per axis \\
& Third person camera rotation & $\pm 2^\circ$ per axis \\
& Third person camera focal length & $\pm 2.0$ \\
& Wrist camera position & $\pm 1$ cm \\
& Wrist camera rotation & $\pm 1^\circ$ \\
& Wrist camera camera focal length & $\pm 1.0$ \\

\midrule
\multirow{6}{*}{Visual (Appearance)}
& Texture vs. color & texture w.p. $0.5$, color w.p. $0.5$ \\
& Texture map & $U(\texttt{texture\_maps})$ (if texture) \\
& Diffuse color & $U(0,1)^3$ (if color) \\
& Texture scale & $U(0.7, 5.0)$ \\
& Roughness & $U(0.0, 1.0)$ \\
& Metallic & $U(0.0, 1.0)$ \\
& Specular & $U(0.0, 1.0)$ \\

\midrule
\multirow{3}{*}{Lighting}
& HDRI texture map & $U(\texttt{texture\_maps})$ \\
& HDRI rotation & $U(\mathrm{SO(3)})$ \\
& HDRI intensity & $U(1000, 4000)$ \\

\midrule
\multirow{4}{*}{Image Augmentation}
& Color jitter 
& brightness $=0.1$, contrast $=0.1$, saturation $=0.1$, hue $=0.05$ \\

& Gaussian blur 
& kernel size $=5$, $\sigma \sim U(0.1, 2.0)$ \\

& Random grayscale 
& $p = 0.05$ \\

& Gaussian noise 
& std $=0.01$ \\

\bottomrule
\end{tabular}
\caption{Domain randomization and state initialization used during training. 
Parameters are either sampled directly from distributions or applied as multiplicative 
scaling to nominal values (denoted by $\times$).}
\label{tab:randomization}
\end{table}

\subsubsection{Randomizations}
\label{sec:randomizations}
We apply domain randomization across dynamics, actuation, state initialization, and visual observations, as summarized in Table~\ref{tab:randomization}. We find that these randomizations significantly improve sim-to-real transfer and robustness. 

For camera setup, we first calibrate extrinsics using ArUco markers in the real environment. We then manually refine alignment by adjusting camera pose and field-of-view (FOV) in simulation to match overlaid real and simulated images. To account for residual calibration error and potential perturbations such as camera movement, we randomize camera pose and focal length within small ranges around the calibrated values. 

For visual observations, we follow extensive domain randomization similar to DextrAH-RGB~\citep{singh2025dextrahrgbvisuomotorpoliciesgrasp}, varying lighting, backgrounds, object and robot appearance, and workspace textures. We randomize between 920 high-dynamic range conditions (HDR) for lighting and 957 texture paths for object and gripper fingertip textures. In addition, we apply standard image augmentations including color jitter, blur, and grayscale. 

We note that the ranges of randomization were not extensively tuned, and were chosen to provide broad coverage of plausible variations. Future work could study how to more systematically select these ranges, either by narrowing them to improve data 
efficiency and alignment with the real system, or by expanding them further to increase diversity and potentially improve robustness.

\subsubsection{Camera placement}
We find that camera placement is critical for successful RGB-based distillation. In particular, using a wrist-mounted camera together with third-person cameras positioned close to the workspace along the $z$-axis provides significantly better performance than top-down viewpoints.

We use two third-person cameras: one capturing the entire tabletop region for grasping, and another focused on the insertion area for contact-rich interactions. This setup helps reduce partial observability by ensuring that both pre-contact and contact phases of the task are well observed.

In practice, we find that a single third-person camera can already achieve reasonable performance. We choose to use both third-person cameras to maximize performance.

\subsubsection{RL Training and Finetuning}
We train RL policies using an asymmetric actor-critic setup ~\citep{pinto2017asymmetric}, where the policy observes object and robot positions, while the critic has access to privileged information such as object mass and other physical parameters. The policy input consists of a stack of 5 frames at 10 Hz to provide short-term temporal context, whereas the critic does not use any history. We find that this design significantly improves training stability under large-scale domain randomization.

We find that directly training RL policies with system-identified dynamics, particularly high joint friction, leads to poor learning performance. In practice, the policy struggles to explore effectively and training often fails to converge.

To address this, we adopt a staged training procedure. We first train the policy to convergence under simplified dynamics, without system-identified parameters (with zero friction, armature, and motor delay), which enables efficient exploration and stable learning. We then apply a curriculum that gradually transitions the environment dynamics toward the system-identified parameters, conditioned on task success. In conjunction with this transition, we also increase controller gains to ensure that the robot remains responsive under higher friction. Note that this is not a curriculum over behaviors or state distributions, since the policy is already succeeding, but rather a parameter curriculum designed specifically to facilitate stable sim-to-real transfer.

Training the RL policy from scratch takes approximately 8 hours for 
\texttt{Peg Insertion}, 16 hours for \texttt{Leg Twisting}, and 24 hours for \texttt{Drawer Assembly} using 4 L40S GPUs. Fine-tuning with the curriculum takes approximately 8 hours on 1 L40S GPU for \texttt{Peg Insertion}, 24 hours on 4 L40S GPUs for \texttt{Leg Twisting}, and 24 hours on 1 L40S GPU for \texttt{Drawer Assembly}.

\subsubsection{Seed Selection}

\begin{wrapfigure}{r}{0.6\linewidth}
    \vspace{-0.5cm}
    \centering
    \begin{subfigure}{\linewidth}
        \includegraphics[width=\linewidth]{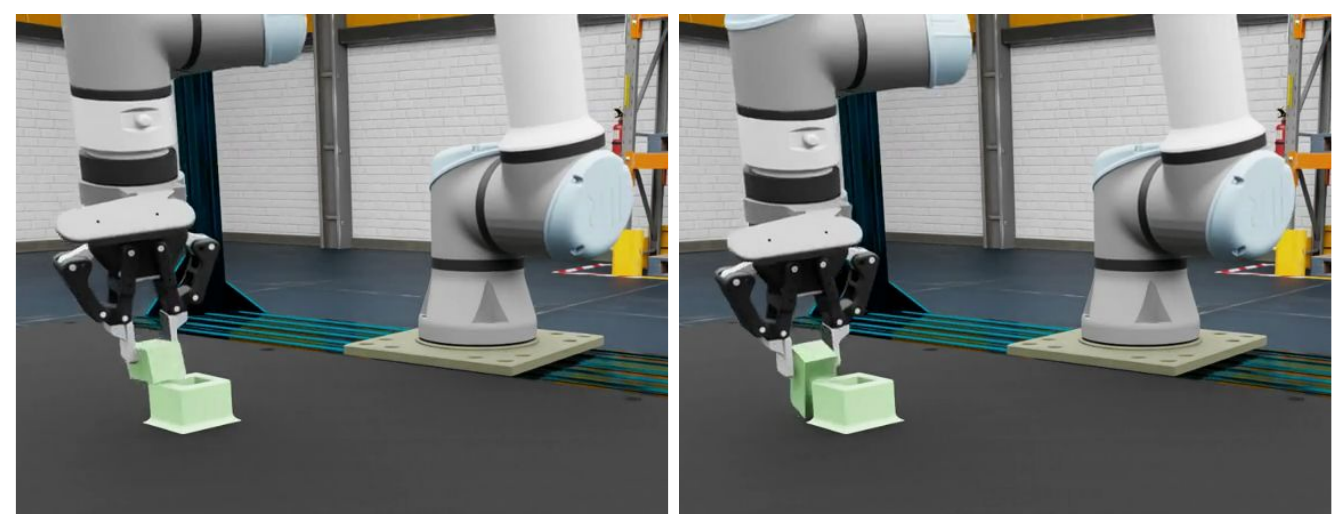}
        \caption{\footnotesize
        \textbf{Behavior which doesn't transfer well.} We show an example of a behavior which doesn't transfer well, where the policy reorient the peg by dropping it above the hole and re-grasping it.}
        \label{fig:untransferable}
    \end{subfigure} 
    \vspace{-0.5cm}
\end{wrapfigure}

We observe that policies trained with different random seeds can exhibit 
substantially different behaviors, with some transferring more reliably to the real world than others. For example, in the peg insertion task, certain policies learn to reorient the peg by dropping it above the hole and re-grasping it as shown in Fig.~\ref{fig:untransferable}. While effective in simulation, this strategy requires precise timing and accurate gripper behavior, making it sensitive to sim-to-real discrepancies and prone to failure on hardware.

To mitigate this, we train multiple policies with different random seeds and select among them using an offline proxy metric. Specifically, we evaluate each policy under injected action noise and measure its success rate. We then select the policy with the highest performance under noise. This metric serves as a proxy for robustness to control error and sim-to-real mismatch, and we find that it correlates well with real-world performance.

\subsubsection{Student-Teacher Distillation}

We collect 80K trajectories from a state-based expert under extensive randomization (Section~\ref{sec:randomizations}) and use them to train a visuomotor policy. The policy consists of an ImageNet-pretrained ResNet-18 encoder followed by a 4-layer MLP with 512 hidden units, trained via supervised learning. The policy takes as input a stack of 5 consecutive frames and predicts the mean and standard deviation of a Gaussian action distribution.

Following DextrAH-RGB~\citep{singh2025dextrahrgbvisuomotorpoliciesgrasp}, we find that combining pose reconstruction with KL matching to the expert’s action distribution leads to the best distillation and transfer performance. In particular, our most performant policies are trained with a KL loss and deployed only using the mean action. Without pose reconstruction, the RGB policy sometimes fails to reliably localize and grasp objects. We also find that end-to-end training, with gradients propagated through the visual encoder, is critical; freezing the encoder significantly degrades performance.

We experiment with alternative representations and architectures, including DINO features and fine-tuning larger pretrained models such as Pi-0.5, but do not observe significant performance improvements. These approaches substantially increase training cost, suggesting that larger models may only provide benefits at larger data scales or across more diverse tasks.

We also evaluate action chunking and diffusion-based policies, but find that they underperform simple MLP policies. We hypothesize that this is due to the high-frequency and reactive nature of the expert RL policy, particularly in contact-rich assembly tasks where small corrective motions and jittering are important.

In terms of data scaling, we observe that using 10K--80K trajectories yields similar performance in simulation, but real-world transfer improves significantly with larger datasets. Training for a sufficient number of iterations is also important; we train for approximately 350K iterations (about 2 days on a single H200 GPU). Collecting 80K trajectories requires approximately 24 GPU hours on a 3090 GPU. Further study is needed to better understand the relationship between 
dataset size, training time, and transfer performance.

Despite these efforts, RGB policy performance in simulation remains limited (approximately 50\% success rate), indicating a gap between imitation learning and robust deployment. This suggests that incorporating online adaptation methods in simulation, such as image-based DAgger or RL fine-tuning, may be necessary to achieve higher performance. We leave this to future work.

\subsubsection{Inference Speed}
We deploy the policy at 10\,Hz using a non-blocking control pipeline. Sensor data is collected asynchronously at their native frequencies. At each control step, the policy retrieves the most recent observation from each sensor that precedes the current timestep, performs inference, and sends the resulting action to the robot.

This design enables smooth and responsive motion, but introduces latency due to sensor delay, data processing, and policy inference. Our setup uses a RealSense D415 wrist camera, along with D435 and D455 third-person cameras, all streamed over USB 3.2. We measure approximately 30\,ms of sensor-to-PC delay and an additional 5\,ms for 
processing. Policy inference using a ResNet18-MLP architecture takes approximately 5\,ms on an NVIDIA RTX 4090 GPU, resulting in a total end-to-end latency of roughly 40\,ms.

Despite not incorporating explicit latency randomization during training, we find that policies remain smooth and robust under this level of delay. However, fast inference speed is critical. When we artificially increase inference latency to 60\,ms, policies exhibit significant jitter and performance degrades substantially.

These results highlight the importance of low-latency inference for stable control and suggest that robustness to delay remains an important direction for future work.

\subsection{Real World Evaluation Details}
\label{sec:appendix_evaluation}

\paragraph{Evaluation Setup.}
We deploy the policy at 10 Hz, matching the control frequency used during simulation training. At the start of evaluation, the robot joints are reset to a fixed initial configuration; no further resets are performed, and the policy continuously rolls out in the environment.

To improve robustness, we implement a simple stuck-detection mechanism. If the robot’s joint positions exhibit negligible movement over a 2-second window (maximum joint displacement below 0.002 radians across all joints), we override the policy by commanding the gripper to open for 1 second. This helps the robot recover from situations such as failed grasps or unfavorable contact configurations. Triggering this mechanism is not counted as a failure in our evaluation.

With this mechanism, the robot operates continuously without manual resets. We define a trajectory as a failure only if human intervention is required, such as repositioning the robot or the object. This typically occurs when the robot becomes stuck or takes excessively long to complete the task.

\paragraph{Initialization Distribution.}
For evaluation, we sample initial object configurations by uniformly randomizing their positions on the table. We intentionally include challenging setups, such as objects in contact that require separation, stacked objects, or objects placed far from the workspace center, for example near the robot base, to stress test the policy’s capabilities.

\paragraph{Physical Setup.}
Task success also depends on aspects of the physical setup. We use a compliant silicone mat on the workspace, which facilitates manipulation by allowing slight deformation during contact, particularly for reorientation and flipping behaviors. The mat is secured to the table using command strips to prevent motion.

For the \texttt{Peg Insertion} and \texttt{Leg Twisting} tasks, we additionally fix the receptive objects, such as the peg hole and tabletop, to the workspace using command strips. This is important because, in simulation, these objects are static, and the policy learns to reorient the insertive object relative to a fixed reference. Without securing them, the receptive objects may move during interaction, leading to failures. If the robot applies sufficient force to dislodge these fixtures during evaluation, we exclude the corresponding trajectory from success rate calculations.

For the \texttt{Drawer Assembly} task, directly fixing the drawer is insufficient due to larger interaction forces. Instead, we secure an L-shaped 3D-printed bracket behind the drawer box, following \citep{heo2023furniturebench}, which provides a more stable constraint. We find that the policy largely ignores this additional structure, likely due to domain randomization during training. Additionally, to address height misalignment caused by the drawer lip, we use layered mats with a cutout to align the drawer bottom with the workspace surface, which significantly improves insertion success. Reducing reliance on such environment modifications is an important direction for future work, for example through the use of bimanual manipulation.

For all tasks, we mount a stage light above the workspace to ensure consistent lighting conditions throughout the day.

\paragraph{Metrics and Evaluation Duration.}
We continuously run the policy for approximately 25 minutes for the peg and drawer tasks, and 50 minutes for the leg task, collecting 41, 52, and 55 trajectories, respectively.

We define a first-try success as completing the task without re-grasping or re-inserting the object (e.g., dropping the object and attempting again counts as a retry). Throughput is computed as the total number of successful trajectories divided by the total evaluation time.

\end{document}